\documentclass{article}

\PassOptionsToPackage{numbers, compress}{natbib}
\usepackage[preprint]{neurips_2026}

\usepackage[utf8]{inputenc}
\usepackage[T1]{fontenc}
\usepackage{lmodern}
\usepackage{hyperref}
\usepackage{url}
\usepackage{booktabs}
\usepackage{amsfonts}
\usepackage{amssymb}
\usepackage{amsmath}
\usepackage{nicefrac}
\usepackage{microtype}
\usepackage{xcolor}
\usepackage{graphicx}
\usepackage{enumitem}
\usepackage[section]{placeins}

\definecolor{accent}{HTML}{1f4e79}
\hypersetup{colorlinks=true, linkcolor=accent, citecolor=accent, urlcolor=accent,
  pdftitle={Probable but Not Valid: A Psychometric Audit of Decision Models as Synthetic Survey Respondents},
  pdfauthor={Mantas Lukauskas, Viktorija Sarkauskaite},
  pdfkeywords={synthetic respondents, silicon samples, LLM survey simulation, psychometrics, decision models, distributional alignment, proper scoring rules, ranked probability score, Gaussian copula, model versions}}

\newcommand{\vDCopulaOosMed}{$-$0.21}
\newcommand{\vDCopulaOosMedLo}{$-$0.39}
\newcommand{\vDCopulaOosMedHi}{$-$0.03}
\newcommand{\vDCopulaOosDemo}{$-$0.11}
\newcommand{\vDCopulaOosDemoLo}{$-$0.35}
\newcommand{\vDCopulaOosDemoHi}{+0.13}

\newcommand{\vGptMedSE}{0.06}
\newcommand{\vJevMedSE}{0.17}
\newcommand{\vNDraws}{200}
\newcommand{\vGptTotal}{0.743}

\newcommand{\vGptMed}{0.929}

\newcommand{\vCopulaTotal}{0.927}

\newcommand{\vCopulaMed}{0.968}
\newcommand{\vCopulaOosTotal}{0.824}
\newcommand{\vCopulaOosCorr}{0.805}
\newcommand{\vCopulaOosMed}{0.722}
\newcommand{\vCopulaOosDemo}{0.382}

\newcommand{\vCopulaDemo}{0.663}

\newcommand{\vGptDemo}{0.489}

\newcommand{\vOracleTotal}{0.886}
\newcommand{\vOracleCorr}{0.923}
\newcommand{\vOracleMed}{0.707}
\newcommand{\vJevTotal}{0.681}

\newcommand{\vJevMed}{0.498}
\newcommand{\vJevArgTotal}{0.786}
\newcommand{\vJevArgMed}{0.972}

\newcommand{\vMercurySampTotal}{0.761}
\newcommand{\vMercurySampCorr}{0.463}

\newcommand{\vLunaTotal}{0.547}
\newcommand{\vLunaCorr}{0.339}

\newcommand{\vGptRawTotal}{0.703}
\newcommand{\vGptAcqFair}{+0.21}
\newcommand{\vGptAcqPublishedStyle}{+1.28}
\newcommand{\vHumanExtreme}{0.270}
\newcommand{\vGptExtreme}{0.024}
\newcommand{\vHumanKeying}{2.001}
\newcommand{\vGptKeying}{1.512}
\newcommand{\vDecAcqMin}{$-$0.27}
\newcommand{\vDecAcqMax}{+3.54}
\newcommand{\vDecExtremeMax}{0.196}
\newcommand{\vHumanPerson}{0.805}

\newcommand{\vDecPersonMin}{0.145}
\newcommand{\vDecPersonMax}{0.495}
\newcommand{\vDecBetweenMin}{0.024}
\newcommand{\vDecBetweenMax}{0.304}
\newcommand{\vDecJsMixMin}{0.037}
\newcommand{\vDecJsMixMax}{0.156}
\newcommand{\vDecJsRoundMin}{0.218}
\newcommand{\vDecJsRoundMax}{0.411}
\newcommand{\vHumanRtwo}{0.28}
\newcommand{\vGptRtwo}{$-$1.79}
\newcommand{\vJevRtwo}{0.30}
\newcommand{\vCopulaRtwo}{0.29}
\newcommand{\vWeightRho}{0.95}
\newcommand{\vWeightRhoMin}{0.71}
\newcommand{\vWeightFrac}{0.60}

\newcommand{\vWeightBottomFrac}{0.48}
\newcommand{\vGptAttPerf}{0.748}
\newcommand{\vHumanAttPerf}{0.372}
\newcommand{\vGptEngPerf}{0.567}
\newcommand{\vHumanEngPerf}{0.475}
\newcommand{\vSimDecDec}{0.384}
\newcommand{\vSimDecHuman}{0.422}

\newcommand{\vNDecision}{14}
\newcommand{\vNDecisionWord}{fourteen}
\newcommand{\vNProvidersWord}{eleven}
\newcommand{\vTotalCost}{1.27}
\newcommand{\vBestGenName}{MiniMax M3}
\newcommand{\vSecondGenName}{MiniMax M2.7}
\newcommand{\vThirdGenName}{Kimi K2.7 Code}
\newcommand{\vBestNewName}{Claude Sonnet 5.5}
\newcommand{\vWorstNewName}{GPT-6.1 Sol}
\newcommand{\vBestNoMedName}{MiniMax M2.7}
\newcommand{\vBestCorrName}{MiniMax M2.7}
\newcommand{\vNGen}{47}
\newcommand{\vNGenFirst}{36}
\newcommand{\vNGenNew}{11}
\newcommand{\vGptRankFirst}{10}

\newcommand{\vBestNewRank}{15}
\newcommand{\vLunaGenRank}{40}
\newcommand{\vGenRtwoPos}{25}
\newcommand{\vGenRtwoNear}{6}
\newcommand{\vGenNullMin}{0}
\newcommand{\vGenNullMax}{6}
\newcommand{\vGenMedNull}{2}

\newcommand{\vSuccDown}{8}
\newcommand{\vSuccLevel}{1}
\newcommand{\vSuccUp}{1}
\newcommand{\vSuccDownNoMed}{9}
\newcommand{\vSuccN}{10}
\newcommand{\vGenAboveDecCorr}{33}
\newcommand{\vJevArgRankGen}{5}
\newcommand{\vJevRankGen}{27}
\newcommand{\vNWithinBest}{12}
\newcommand{\vGainMediDown}{30}
\newcommand{\vGDecBelow}{13}
\newcommand{\vGDecN}{14}
\newcommand{\vPctAgree}{68}
\newcommand{\vPctN}{77}

\newcommand{\vSuccBelow}{5}

\newcommand{\vSuccSameTierDown}{5}
\newcommand{\vSuccSameTierMove}{6}
\newcommand{\vDecDistAboveGenMed}{2}
\newcommand{\vNHigh}{6}
\newcommand{\vNHighUp}{3}
\newcommand{\vWaveNTests}{10}
\newcommand{\vBestGenTotal}{0.813}

\newcommand{\vBestGenMed}{0.998}

\newcommand{\vSecondGenTotal}{0.809}
\newcommand{\vThirdGenTotal}{0.802}
\newcommand{\vGenFirstMedTotal}{0.701}
\newcommand{\vGenNewMedTotal}{0.641}
\newcommand{\vBestNewTotal}{0.735}
\newcommand{\vWorstNewTotal}{0.510}
\newcommand{\vGenMedTotal}{0.695}

\newcommand{\vGenMedDist}{0.749}

\newcommand{\vGenCorrMax}{0.807}
\newcommand{\vGenDistMin}{0.618}
\newcommand{\vGenDistMax}{0.903}
\newcommand{\vLunaGenTotal}{0.621}
\newcommand{\vLunaGenCorr}{0.566}
\newcommand{\vLunaGenDist}{0.686}

\newcommand{\vLunaArgTotal}{0.568}
\newcommand{\vLunaDist}{0.676}

\newcommand{\vOpusFiveTotal}{0.606}
\newcommand{\vOpusFourSevenTotal}{0.789}
\newcommand{\vOpusFourEightTotal}{0.740}
\newcommand{\vGenFirstMedNoMed}{0.765}
\newcommand{\vGenNewMedNoMed}{0.734}
\newcommand{\vBestNoMedTotal}{0.821}
\newcommand{\vKimiCodeTotal}{0.802}
\newcommand{\vKimiThreeTotal}{0.641}
\newcommand{\vDecCorrMax}{0.667}

\newcommand{\vMedGain}{0.093}
\newcommand{\vMinGain}{0.016}
\newcommand{\vRankRho}{0.79}
\newcommand{\vGenNewP}{0.06}
\newcommand{\vGenNewPNoMed}{0.09}
\newcommand{\vSuccSignP}{0.04}
\newcommand{\vGenMedPerson}{0.53}
\newcommand{\vGenMedRtwo}{0.10}
\newcommand{\vGenMedAcq}{0.21}
\newcommand{\vGainKeyRho}{0.21}
\newcommand{\vGainKeyP}{0.23}
\newcommand{\vLunaGenPerson}{0.40}
\newcommand{\vLunaPerson}{0.36}
\newcommand{\vLunaGenAcq}{0.53}
\newcommand{\vLunaAcq}{1.48}
\newcommand{\vLunaGenRtwo}{$-$0.26}
\newcommand{\vLunaRtwo}{0.18}
\newcommand{\vGenRtwoMax}{0.30}
\newcommand{\vGainCorrSd}{0.07}
\newcommand{\vGainRelSd}{0.18}
\newcommand{\vGainMediSd}{0.19}
\newcommand{\vSuccSameTierP}{0.22}
\newcommand{\vWaveFirstSD}{0.56}
\newcommand{\vWaveNewSD}{0.46}
\newcommand{\vWavePSD}{0.01}
\newcommand{\vWaveFirstExtreme}{0.04}
\newcommand{\vWaveNewExtreme}{0.02}
\newcommand{\vWavePExtreme}{0.09}

\newcommand{\vWaveFirstPerson}{0.55}
\newcommand{\vWaveNewPerson}{0.47}
\newcommand{\vWavePPerson}{0.05}
\newcommand{\vWaveFirstRtwo}{$-$0.06}
\newcommand{\vWaveNewRtwo}{0.15}
\newcommand{\vWavePRtwo}{0.11}
\newcommand{\vWaveFirstDemo}{0.45}
\newcommand{\vWaveNewDemo}{0.37}
\newcommand{\vWavePDemo}{0.02}
\newcommand{\vWaveFirstCorr}{0.73}
\newcommand{\vWaveNewCorr}{0.70}

\newcommand{\vWaveFirstDist}{0.76}
\newcommand{\vWaveNewDist}{0.72}

\newcommand{\vWaveHolmMin}{0.14}
\newcommand{\vGainCorrMed}{+0.37}
\newcommand{\vGainRelMed}{+0.11}
\newcommand{\vGainMediMed}{$-$0.18}
\newcommand{\vBCopulaOosTotal}{+0.01}
\newcommand{\vBCopulaOosTotalLo}{$-$0.07}
\newcommand{\vBCopulaOosTotalHi}{+0.09}
\newcommand{\vBCopulaTotal}{+0.11}
\newcommand{\vBCopulaTotalLo}{+0.06}
\newcommand{\vBCopulaTotalHi}{+0.17}
\newcommand{\vBOracleTotal}{+0.07}
\newcommand{\vBOracleTotalLo}{0.00}
\newcommand{\vBOracleTotalHi}{+0.15}
\newcommand{\vBJevArgTotal}{$-$0.03}
\newcommand{\vBJevArgTotalLo}{$-$0.10}
\newcommand{\vBJevArgTotalHi}{+0.04}
\newcommand{\vBJevArgCorr}{$-$0.10}
\newcommand{\vBJevArgCorrLo}{$-$0.17}
\newcommand{\vBJevArgCorrHi}{$-$0.03}
\newcommand{\vBJevTotal}{$-$0.13}
\newcommand{\vBJevTotalLo}{$-$0.21}
\newcommand{\vBJevTotalHi}{$-$0.05}
\newcommand{\vBGptTotal}{$-$0.07}
\newcommand{\vBGptTotalLo}{$-$0.13}
\newcommand{\vBGptTotalHi}{$-$0.01}
\newcommand{\vBGptCorr}{$-$0.16}
\newcommand{\vBGptCorrLo}{$-$0.24}
\newcommand{\vBGptCorrHi}{$-$0.09}

\newcommand{\vLunaPair}{+0.07}
\newcommand{\vLunaPairLo}{0.00}
\newcommand{\vLunaPairHi}{+0.15}
\newcommand{\vLunaPairCorr}{+0.23}
\newcommand{\vLunaPairCorrLo}{+0.12}
\newcommand{\vLunaPairCorrHi}{+0.33}

\newcommand{\vLunaPairArgCorr}{+0.18}
\newcommand{\vLunaPairArgCorrLo}{+0.07}
\newcommand{\vLunaPairArgCorrHi}{+0.29}
\newcommand{\vSuccMedDelta}{$-$0.08}
\newcommand{\vGJevArg}{+0.04}
\newcommand{\vGJevArgLo}{$-$0.02}
\newcommand{\vGJevArgHi}{+0.10}

\newcommand{\vCtrlCost}{4.33}

\newcommand{\vReplMin}{0.82}
\newcommand{\vRetestCompRMin}{0.95}
\newcommand{\vJevRetestSame}{95}
\newcommand{\vDOneRetestSame}{92}
\newcommand{\vJevCzeroUnique}{99}
\newcommand{\vJevCzeroNoise}{0.02}
\newcommand{\vMsRetestSame}{95}
\newcommand{\vSolarRetestSame}{89}
\newcommand{\vSolarCzeroUnique}{100}
\newcommand{\vSolarCzeroNoise}{0.52}
\newcommand{\vShiftMin}{0.04}
\newcommand{\vShiftMax}{0.18}

\newcommand{\vCzeroRoundMin}{0.11}
\newcommand{\vCzeroRoundMax}{0.37}
\newcommand{\vCzeroSampMin}{0.39}
\newcommand{\vCzeroSampMax}{0.43}
\newcommand{\vGainMin}{0.06}
\newcommand{\vGainMax}{0.33}
\newcommand{\vGainMed}{0.22}
\newcommand{\vGenGainMed}{0.16}

\newcommand{\vNCzero}{13}
\newcommand{\vJsWorse}{12}
\newcommand{\vTevFormatSame}{98}
\newcommand{\vTevFormatTv}{0.012}
\newcommand{\vEduShiftMin}{0.12}
\newcommand{\vEduShiftMax}{0.37}
\newcommand{\vEduShiftMed}{0.22}
\newcommand{\vGenEduShiftMed}{0.32}

\newcommand{\vEduDMed}{1.06}
\newcommand{\vGenEduDMed}{0.68}
\newcommand{\vHumanEduD}{0.44}
\newcommand{\vEduSignDec}{93}
\newcommand{\vEduSignGen}{100}

\newcommand{\vRoleShiftMin}{0.12}
\newcommand{\vRoleShiftMax}{0.62}
\newcommand{\vRoleShiftMed}{0.22}
\newcommand{\vGenRoleShiftMed}{0.47}

\newcommand{\vHumanRoleD}{0.46}
\newcommand{\vRoleSignDec}{96}
\newcommand{\vRoleSignGen}{100}
\newcommand{\vNRoleDec}{13}
\newcommand{\vGenderShiftMin}{0.01}
\newcommand{\vGenderShiftMax}{0.16}
\newcommand{\vGenderShiftMed}{0.03}
\newcommand{\vGenGenderShiftMed}{0.06}

\newcommand{\vGenderDMed}{0.31}
\newcommand{\vGenGenderDMed}{0.17}

\newcommand{\vGenderSignDec}{47}
\newcommand{\vGenderSignGen}{50}

\newcommand{\vNGenEdu}{37}
\newcommand{\vRoleOverEdu}{6}
\newcommand{\vNRoleEdu}{13}
\newcommand{\vRoleBelowGen}{5}
\newcommand{\vRoleBelowHalf}{7}

\newcommand{\vHumanEduDLo}{0.20}
\newcommand{\vHumanEduDHi}{1.04}
\newcommand{\vRoleShiftMedLo}{0.20}
\newcommand{\vRoleShiftMedHi}{0.25}
\newcommand{\vHumanRoleDLo}{0.31}
\newcommand{\vHumanRoleDHi}{0.62}

\newcommand{\vFcRegGain}{1.0}
\newcommand{\vFcRegLo}{0.1}
\newcommand{\vFcRegHi}{1.8}
\newcommand{\vFcUniWorse}{42}

\newcommand{\vFcDecWorseMin}{6}
\newcommand{\vFcDecWorseMax}{55}
\newcommand{\vFcDecWorseMed}{20}
\newcommand{\vFcDecLogWorseMin}{7}
\newcommand{\vFcDecLogWorseMax}{62}
\newcommand{\vFcDecLogNeg}{14}
\newcommand{\vFcPoolWorseMin}{7}
\newcommand{\vFcPoolWorseMax}{51}

\newcommand{\vFcNDec}{14}
\newcommand{\vFcRoundWorseMin}{48}
\newcommand{\vFcRoundWorseMax}{105}

\newcommand{\vFcCthreeBetter}{11}
\newcommand{\vFcNCzero}{13}
\newcommand{\vFcPoolBetter}{7}
\newcommand{\vFcPoolDiffMed}{0.3}

\newcommand{\vFcGenWorseMed}{57}
\newcommand{\vFcGenWorseMin}{41}
\newcommand{\vFcGenWorseMax}{104}
\newcommand{\vFcGenDrawWorseMed}{78}

\newcommand{\vFcGenPoolBetter}{31}
\newcommand{\vFcGenPoolDiffMed}{27}
\newcommand{\vFcDecEntMed}{1.37}
\newcommand{\vFcGenEntMed}{0.32}
\newcommand{\vFcPopEnt}{1.36}

\newcommand{\vFcShrinkMin}{$-$0.5}
\newcommand{\vFcShrinkMax}{0.9}
\newcommand{\vFcShrinkLamMin}{0.05}
\newcommand{\vFcShrinkLamMax}{0.25}

\newcommand{\vFcGenShrinkMed}{0.7}
\newcommand{\vFcGenShrinkMin}{$-$0.3}
\newcommand{\vFcGenShrinkMax}{1.8}

\newcommand{\vFcDrawCostMed}{5}
\newcommand{\vFcDrawWorseMin}{12}
\newcommand{\vFcDrawWorseMax}{64}
\newcommand{\vDecRhoMin}{0.10}
\newcommand{\vDecRhoMax}{0.19}
\newcommand{\vGenRhoMin}{0.10}
\newcommand{\vGenRhoMax}{0.20}

\newcommand{\vRegRho}{0.15}
\newcommand{\vRegRhoLo}{0.08}
\newcommand{\vRegRhoHi}{0.22}
\newcommand{\vDecCompRMin}{0.21}
\newcommand{\vDecCompRMax}{0.36}

\newcommand{\vRegCompR}{0.24}
\newcommand{\vRegBetween}{0.012}
\newcommand{\vDecBetweenRatioMin}{2}
\newcommand{\vDecBetweenRatioMax}{25}

\newcommand{\vRegRCorr}{0.23}
\newcommand{\vRegACorr}{0.14}
\newcommand{\vRegSCorr}{0.03}
\newcommand{\vDecAboveRegCorr}{12}

\newcommand{\vNCopSeeds}{20}
\newcommand{\vRefCopOos}{0.71}

\newcommand{\vRefCopSeedMean}{0.75}
\newcommand{\vRefRepMean}{0.79}
\newcommand{\vRefRepMin}{0.76}
\newcommand{\vRefRepMax}{0.84}
\newcommand{\vRefRepCorr}{0.73}
\newcommand{\vRefRepDist}{0.99}
\newcommand{\vRefRepMed}{0.56}
\newcommand{\vRefRepDemo}{0.53}
\newcommand{\vRefBestGen}{0.84}
\newcommand{\vRefGenAboveRep}{8}
\newcommand{\vRefGenCorrMax}{0.78}

\newcommand{\vRefGenDistMax}{0.90}

\newcommand{\vRefBigCopCorr}{0.77}
\newcommand{\vRefBigCopDist}{0.98}

\newcommand{\vRefBestGenName}{MiniMax M2.7}
\newcommand{\vStrHumanR}{0.22}
\newcommand{\vStrOracleR}{0.25}
\newcommand{\vStrGenRMed}{0.36}
\newcommand{\vStrDecRMin}{0.13}
\newcommand{\vStrDecRMax}{0.44}
\newcommand{\vStrDecRMed}{0.28}
\newcommand{\vStrHumanEig}{0.28}
\newcommand{\vStrOracleEig}{0.33}
\newcommand{\vStrGenEigMed}{0.49}
\newcommand{\vStrHumanWB}{0.16}

\newcommand{\vStrDecWBMin}{0.02}

\newcommand{\vStrHumanAlpha}{0.81}
\newcommand{\vStrGenAlphaMed}{0.88}

\newcommand{\vStrOraclePhi}{0.97}
\newcommand{\vStrGenPhiMed}{0.82}

\newcommand{\vStrDecPhiMin}{0.39}
\newcommand{\vStrDecPhiMax}{0.79}
\newcommand{\vStrDecEigExpMin}{0.52}
\newcommand{\vStrDecEigExpMax}{0.85}
\newcommand{\vDecSdRatioMin}{0.24}
\newcommand{\vDecSdRatioMax}{0.59}
\newcommand{\vStrDecWBBelow}{13}

\newcommand{\vStrGenRAbove}{42}
\newcommand{\vNullDrawn}{10}
\newcommand{\vNullPartWhole}{3}
\newcommand{\vNullN}{7}
\newcommand{\vGptNull}{6}

\newcommand{\vDecNullMin}{0}
\newcommand{\vDecNullMax}{6}

\newcommand{\vGenNullZero}{6}
\newcommand{\vDeciderOldTotal}{0.659}
\newcommand{\vDeciderNewTotal}{0.660}
\newcommand{\vDDeciderTotal}{0.00}
\newcommand{\vDDeciderTotalLo}{$-$0.05}
\newcommand{\vDDeciderTotalHi}{+0.05}

\newcommand{\vDDeciderCorr}{$-$0.01}
\newcommand{\vDDeciderCorrLo}{$-$0.03}
\newcommand{\vDDeciderCorrHi}{+0.02}

\title{Probable but Not Valid:\\A Psychometric Audit of Decision Models\\as Synthetic Survey Respondents}

\author{%
  Mantas Lukauskas\\
  Hostinger; Kaunas University of Technology;\\
  AI Insight Lab\\
  \texttt{mantas.lukauskas@ktu.lt}
  \And
  Viktorija \v{S}arkauskait\.{e}\\
  AI Insight Lab;\\
  Independent Researcher\\
  \texttt{viktorija@aiinsightlab.ai}
}

\begin{document}
\maketitle

\begin{abstract}
Language models are increasingly used to create synthetic survey samples, also called silicon samples. Decision models are a new class of language models that return a probability for each point of a rating scale instead of text. We ask whether these models, or the newest generative LLMs, can stand in for human respondents on a psychometric survey. We test \vNDecisionWord{} decision models from \vNProvidersWord{} providers and \vNGen{} generative LLMs on an open Lithuanian organisational-psychology survey ($n{=}263$, 68 items), with 100 real respondent profiles, the Psychometric Similarity Score (PSS) and statistical baselines. The best generative model, \vBestGenName{}, reaches PSS \vBestGenTotal{}. This is level with a Gaussian copula fit only on other respondents (\vCopulaOosTotal{}) and below a copula fit in sample (\vCopulaTotal{}). A second sample of 100 real people scores below the best models, so at this size PSS cannot rank them against real respondents. In \vSuccBelow{} of \vSuccN{} product lines the newest generative release scores below an earlier release of the same line beyond resampling noise, and none gains beyond it. With the default rounded readout, \vGDecBelow{} of \vGDecN{} decision models score below the generative reference gpt-5.4-mini. Of the \vNGen{} generative models, \vGenAboveDecCorr{} reproduce the pattern of item correlations better than any decision model. Averaged over respondents, the probability vectors are much closer to the human item marginals than rounded answers. They also rank respondents about as well as a profile regression fit on 163 real people. As forecasts of each person's own answers, however, every decision model is worse than the answer frequencies of other respondents, mainly because its average vector differs from them. The profile regression improves on these frequencies by only \vFcRegGain\%, so the profile fields say little about how a person answers. Synthetic respondents built on such profiles should therefore aim at sample-level fidelity.
\end{abstract}

\section{Introduction}
\label{sec:intro}
Large language models (LLMs) are now used to simulate survey respondents~\citep{argyle2023outoftheir,horton2023llmpolicies,aher2023using,dillion2023can}, and their role in social science research is widely discussed~\citep{grossmann2023ai,bail2024can,ziems2024can}. Synthetic samples can match human averages and still differ from human data in variance and in regression estimates~\citep{bisbee2024synthetic,boelaert2025machine}. A psychometric audit of 37 LLMs found a similar gap on an open Lithuanian organisational-psychology survey~\citep{lukauskas2026plausible}. The models were conditioned on real respondent profiles, and the synthetic samples were compared with the human sample on item distributions, item correlations, reliability, mediation and demographic effects. The answers looked plausible. However, the samples did not behave like the human sample, and a Gaussian copula fit to the human data did better on the components that describe the sample as a whole. We call this study the original audit below.

In September and October 2026 a new class of models became available through a common API~\citep{openrouter2026decisions}. We call them decision models, which is the name under which the providers list them. The term is unrelated to decision theory. A decision model reads a state (text or JSON) and answers a set of named questions in one request. For a score question with an ordered rubric it returns a probability for every level and the expected level. No text is generated, output tokens are free, and one request can carry 64 to 200 questions. For a survey this looks close to ideal. Each Likert item becomes one score question, the respondent profile becomes the state, and the model returns a full answer distribution for every item. Several failure sources of generated answers cannot occur, such as parsing errors, refusals and long reasoning traces.

The generative models have changed as well. Since the original audit, OpenAI, Anthropic, Google, xAI and several open-weight developers have released new versions. Model behaviour can change between versions in ways that are not announced~\citep{chen2024chatgpt}, so it is an open question whether these newer models close the gap. We ran eleven of them on the same benchmark through vendor APIs and Fireworks. One of them, GPT-6 Luna, is also offered as a decision model. This allows a comparison of the two interfaces for one model name.

This paper aims to test whether decision models are better synthetic respondents than generative LLMs, and whether newer generative LLMs are better than earlier releases. We raised four hypotheses. First, a model that returns a distribution over scale points reproduces the item marginals better than a generative LLM. Second, it does not by itself reproduce the pattern of correlations between items, because the vector for each item is produced without an explicit shared respondent state. Third, it uses the respondent profile only weakly. Fourth, newer generative models score higher than their predecessors. We also ask whether the probability vectors forecast each person's answers better than the answer frequencies of other respondents. All models are compared on the same 100 respondents, profile fields, scoring and baselines.

We use the word \emph{valid} in the sense of the testing standards~\citep{aera2014standards}. Synthetic answers are valid for a use if they support the same interpretations as human answers. We examine two sources of evidence. The first is internal structure: item distributions, item correlations, reliability and factor congruence. The second is relations to other variables: mediation, group contrasts and the prediction of human outcomes. We do not examine content validity.

Our main findings are as follows. Scored against all 263 respondents, the best language models are level with a Gaussian copula fit only on other respondents, and none is above it. Newer generative releases do not score higher. In \vSuccBelow{} of \vSuccN{} product lines the newest release scores below an earlier release of the same line by more than the resampling noise. Averaged over respondents, the vectors of decision models come much closer to the human item marginals than rounded answers do. However, they reproduce the pattern of item correlations and the factor structure less well than most generative LLMs. Their vectors carry about as much information about the person as a regression on the profile, but they are poorly calibrated. As forecasts of each person's own answers they are worse than the answer frequencies of other respondents, and the profile regression improves on these frequencies by only \vFcRegGain\%.

We score every sample with a revised version of the original scoring, which reverses model answers to reverse-keyed items before comparison. Under it the main conclusion of the original audit holds, and the order of the LLMs changes. Appendix~\ref{app:scoring} is a self-contained record of these corrections.

The contributions of this paper are as follows.
\begin{enumerate}[leftmargin=1.4em,itemsep=1pt]
  \item A psychometric evaluation of decision models as synthetic survey respondents: \vNDecisionWord{} models from \vNProvidersWord{} providers, 100 respondents, 68 items and full probability vectors, at a total cost of \$\vTotalCost{}.
  \item An updated generative leaderboard of \vNGen{} LLMs under the revised scoring. It includes \vNGenNew{} models released after the original audit and a comparison of successive releases within \vSuccN{} product lines. In \vSuccBelow{} of these lines the newest release scores lower by more than the resampling noise.
  \item A comparison of the generative and the decision interface of GPT-6 Luna.
  \item An analysis of readouts, of the internal structure of the samples and of what the probability vectors contain. It uses control grids without a profile and with repeated requests, swaps of education, role and gender, and a profile regression that is read out like a decision model.
  \item A test of the probability vectors as forecasts of each person's own answers with a proper scoring rule. For every decision model the vectors are worse than the answer frequencies of other respondents.
  \item A check of the reference points. Besides the real answers of the same 100 people and a copula fit only on other respondents, it uses a second sample of 100 real people and a reference that excludes the training data of the copula.
  \item Nine recommendations for evaluating synthetic survey respondents (Section~\ref{sec:discussion}).
\end{enumerate}

The rest of this paper is organised as follows. Section~\ref{sec:related} reviews related work. Section~\ref{sec:benchmark} summarises the benchmark and data. Section~\ref{sec:method} describes how the models were queried and scored. Section~\ref{sec:results} presents the results, and Section~\ref{sec:discussion} discusses them. Section~\ref{sec:limitations} lists the limitations and directions for future work. Finally, Section~\ref{sec:conclusion} concludes.

\section{Related work}
\label{sec:related}
\paragraph{LLMs as survey respondents.} Persona-conditioned LLMs have been used to reproduce public-opinion splits~\citep{argyle2023outoftheir}, economic choices~\citep{horton2023llmpolicies} and classic experiments~\citep{aher2023using}. Other studies found that LLM opinions reflect a narrow part of the population~\citep{santurkar2023whose,durmus2023towards}, that synthetic samples can match human averages while showing too little variance and different regression estimates~\citep{bisbee2024synthetic}, and that LLMs flatten identity groups~\citep{wang2024llm}. Answers of LLMs to World Values Survey items show low variance and a bias that changes from topic to topic~\citep{boelaert2025machine}. LLM simulations have also predicted the treatment effects of survey experiments about as well as human forecasters~\citep{hewitt2024predicting}. Recent position papers discuss both the promise of LLM social simulations and their limits~\citep{anthis2025llm,kozlowski2025simulating}. In HCI and marketing research, silicon samples have been recommended mainly for ideation and pilot studies~\citep{hamalainen2023evaluating,sarstedt2024using}. \citet{lukauskas2026plausible} moved the comparison to the level of a full psychometric instrument and added statistical baselines and a human ceiling.

\paragraph{Predicting one person.} \citet{park2026llmagents} built agents for 1,052 Americans and scored their answers to held-out General Social Survey items against each person's own test-retest consistency. Agents built from interviews came close to this consistency, and agents given only demographic attributes did worse. \citet{kim2023aiaugmented} fine-tuned a language model on General Social Survey waves. It predicted missing individual answers well, but answers to questions that had never been asked less well. A model fine-tuned on trial-level data from many psychology experiments predicts the choices of held-out participants~\citep{binz2025foundation}. We use a reference that needs no information about the person, namely the answer frequencies of other respondents.

\paragraph{How the answer is elicited.} The format of elicitation changes what an LLM appears to believe. Survey answers of LLMs depend strongly on answer order and labelling, and adjusting for these effects moves the answers toward uniform~\citep{dominguez2024questioning}. Opinion distributions have been read from token probabilities~\citep{santurkar2023whose}, and distributional alignment differs between log-probabilities, sampled answers and stated distributions~\citep{meister2025benchmarking}. For instruction-tuned models, the first-token probabilities over answer options often do not match the answer the model gives in text~\citep{wang2024myanswer}. Fine-tuning on survey data brings the predicted answer distribution closer to the answer distributions of human groups~\citep{cao2025specializing,suh2025language}. Direct numeric ratings give unrealistic rating distributions, and mapping free-text answers onto the scale anchors by semantic similarity gives more realistic ones~\citep{maier2025llms}. All of these studies compare distributions for groups of people. We ask what the distribution says about one person. Decision models make the probability readout the default interface, and the analyst still has to turn each distribution into one answer per respondent.

\paragraph{Model versions.} The behaviour of a deployed model can change between versions in ways that are not announced and not monotone. \citet{chen2024chatgpt} found that GPT-4 and GPT-3.5 changed markedly on several tasks within a few months. Synthetic-respondent studies usually report one version per model. We compare successive releases on the same benchmark.

\paragraph{Psychometrics of LLMs.} Methods from cognitive psychology have been used to study how LLMs decide and reason~\citep{binz2023using}, and psychometric inventories have been applied to the models themselves~\citep{pellert2024ai,serapio2025psychometric}. When models answer such inventories with personas built from the demographics of real people, the psychometric properties of the answers are poor~\citep{petrov2024limited}.

\paragraph{Response styles and baselines.} \citet{tjuatja2024llms} found that LLMs do not show human-like response biases in survey questions, including acquiescence, and that they react to wording changes that do not move human answers. LLMs also shift Big Five answers toward the socially desirable end when they can infer that they are being assessed~\citep{salecha2024large}. We measure acquiescence with the balanced index used in cross-national survey research~\citep{baumgartner2001response}. The statistical anchor is a Gaussian copula with empirical marginals~\citep{nelsen2006introduction}, a standard model for synthetic tabular data~\citep{patki2016synthetic}.

\paragraph{What this paper adds.} Earlier work compares model and human answer distributions for groups of people, mostly on opinion items, and reads probabilities from the next token of a generative model. We make three changes. First, we evaluate models whose only output is a probability vector and compare them with \vNGen{} generative LLMs on the item distributions, item correlations, factor structure, reliability, mediation and group contrasts of four survey scales. Second, we score the vectors as forecasts of each person's own answers with a proper scoring rule, against the answer frequencies of other respondents and a regression on the same profile. Third, we compare successive releases within \vSuccN{} product lines on one fixed benchmark. We are not aware of a published evaluation of decision models on survey data. The model descriptions on the provider pages are the only documentation~\citep{openrouter2026decisions}.

\section{Benchmark and data}
\label{sec:benchmark}
This section summarises the benchmark, which comes from the original audit~\citep{lukauskas2026plausible}.

\paragraph{Data.} The human reference is an open sample of $n{=}263$ Lithuanian employees~\citep{sarkauskaite2020}. They answered the Individual Work Performance Questionnaire (18 items, 1--5)~\citep{koopmans2014}, the Dunham attitudes toward change scale (18 items, 1--5)~\citep{dunham1989}, a change-engagement scale built for the original study (15 items, 1--5) and the UWES-17 work engagement scale (17 items, 1--7)~\citep{schaufeli2003}. The change-engagement scale is less well validated than the other three. Together this gives 68 items in 12 subscales. Thirteen items are reverse-keyed: five in the Dunham scale and eight in the change-engagement scale.

\paragraph{Profiles and sample.} The original audit varied how much of the profile the model sees, from no profile (condition C0) to the full structured profile (C3). We use C3, in which the model sees eleven profile fields: country, gender, age, education, position, total tenure, tenure in the current organisation, sector, organisation size, and two ratings of how intense and how relevant recent organisational change was. The headline sample is 100 respondents stratified by gender and position with a fixed seed. We use exactly the 100 respondent IDs of its gpt-5.4-mini C3 cell.

\paragraph{PSS.} Every synthetic sample is scored against all 263 humans with five components,
\begin{equation}
\mathrm{PSS} = 0.25\,D + 0.25\,C + 0.20\,R + 0.20\,M + 0.10\,G .
\end{equation}
$D$ is one minus the mean Jensen-Shannon divergence (in bits) between the human and synthetic item distributions. $C$ is the Pearson correlation between the upper triangles of the two $68\times68$ item correlation matrices, floored at zero. It compares the pattern of the two matrices and does not penalise a uniform rise of all correlations, so Section~\ref{sec:res-structure} also reports their level and the factor structure. $R$ is one minus the absolute difference in Cronbach's $\alpha$~\citep{cronbach1951} on the Dunham scale. $M$ is one minus the absolute difference in the standardised indirect effect of attitudes on performance through engagement~\citep{hayes2017introduction}, divided by 0.5. $G$ is the share of 96 demographic contrasts (8 contrasts by 12 outcomes) with the same sign of Cohen's $d$ as in humans. It is multiplied by one minus the absolute mean amplification, where the amplification of a contrast is $|d|$ in the synthetic sample minus $|d|$ in humans. Each component lies in $[0,1]$. The weights are those of the original audit.

\paragraph{References.} The original audit compared LLMs with five statistical baselines and a held-out human ceiling, in which 53 humans were scored against the other 210 (PSS 0.825). Its conclusion was that LLM samples are not a drop-in replacement for human survey data.

\paragraph{Scoring.} We use a revised version of the original scoring. It reverses model answers to the 13 reverse-keyed items before they are compared with the reversed human answers, and it includes the distribution component in the totals of the statistical baselines. Appendix~\ref{app:scoring} lists the numbers that change relative to the published version of the original audit.

\section{Querying and scoring the models}
\label{sec:method}
\paragraph{Decision models.} We listed all models with the decisions output modality on OpenRouter on 7 October 2026 and again on 10 October 2026~\citep{openrouter2026decisions}. Appendix Table~\ref{tab:models} gives the pinned versions and costs. We kept every model that accepted five- and seven-level rubrics and at least 64 score questions per request. Three candidates were left out. The Span-01 family accepts only a chat-message state, so the profile cannot be passed in the same form. \texttt{jev-latest} is only an alias that points to Jev~1.13. Solar Decide takes about two minutes per respondent and returned server errors for many requests. After eight hours 46 of 100 respondents were complete, below our threshold of 90, so we left it out. The second listing added five models, and all of them met the criteria. Decider v1.1 27B is described by its provider as the successor of Decider v1 27B, and Solar Decide Flash is a low-latency variant of Solar Decide. Clef Omni, Microsoft-Decision-1 and Drex v1.5 are new. This leaves \vNDecisionWord{} models from \vNProvidersWord{} providers: Jev~1.13, Decider v1 27B, Decider v1.1 27B, Liquid d1, Kev 4B, Tev1 4B, Mercury Decide, Solar Decide Flash, Clef, Clef Flash, Clef Omni, GPT-6 Luna Decisions, Microsoft-Decision-1 and Drex v1.5.

\paragraph{Requests.} The state is the C3 profile as a JSON object with English field names and values. Each item becomes one score question. Its instruction is the Lithuanian sentence ``Answer the statement as this employee would answer it'' followed by the Lithuanian item text, and its criteria are the numbered Lithuanian scale anchors (Appendix~\ref{app:request}). All 68 questions go in one request, in instrument order. The three Clef models and Drex v1.5 reject more than 64 questions, so they receive the same profile twice, first with 64 items and then with the last 4. The endpoint exposes no temperature or seed. Failed requests were retried with exponential backoff, and all 1,400 respondent-model cells completed.

\paragraph{Readouts.} For item $j$ the model returns a vector $p_j$ over $K$ levels. We store the full vector and derive three answers. The rounded expected score (R) is $\mathrm{round}(\sum_k k\,p_{jk})$ on the original scale. The most probable category (A) is $\arg\max_k p_{jk}$, and the sampled category (S) is one draw from $p_j$, repeated five times with a fixed seed and averaged. R is the default readout because it is closest to a single generated answer. The other two readouts show how much the conclusions depend on this choice.

\paragraph{Generative models.} The generative reference set has two parts. First, it contains the C3 cells of the original audit (single call, minimal reasoning) with at least 90 complete respondents among the same 100 IDs. This gives \vNGenFirst{} cells. Qwen 3.7 Plus has 75 complete respondents and is left out of the rankings. Second, it contains \vNGenNew{} models released after the original audit. They are the new releases of the vendors in that audit that we could reach through a vendor API or Fireworks in October 2026. Mistral Large 4 is left out because it is offered only through OpenRouter. We ran these models with the generator, prompt and parser of the original audit and requested a temperature of 0.7. As in that audit, the GPT-5 and GPT-6 families and models whose API rejects the parameter run at the provider default. Each respondent gets one call with all 68 items. We use the same minimal reasoning setting. It means no extended thinking for Anthropic, a thinking budget of zero for Google, and \texttt{reasoning\_effort=minimal} for the OpenAI-compatible APIs. We drop this parameter when an API rejects it. OpenAI, Anthropic and Google models are called through their vendor APIs: GPT-6 Luna and GPT-6.1 Sol, Claude Opus 5.5 and Claude Sonnet 5.5, Gemini 3.6 Flash and Gemini 3.8 Flash. Grok 4.7 is called through the xAI API, and the open-weight Kimi K3, Qwen 3.8 Max, GLM 5.3 and DeepSeek V4.1 Flash run on Fireworks. The original audit reached Grok and the open-weight models through the Nexos gateway. Decision models are only offered through OpenRouter. All new models completed 100 respondents except Grok 4.7 (99), and the run cost \$15.05 at list prices.

\paragraph{What differs between the two classes.} The comparison of decision and generative models changes more than the output format. A generative model answers all 68 items in one generated text, so each answer can depend on the earlier ones. A decision model returns a separate distribution for each item. The decision request uses an English JSON state with Lithuanian items, while the generative prompt is fully Lithuanian. Decision models are reached through OpenRouter, and generative models through vendor APIs and Fireworks. Finally, most decision endpoints are deterministic, while generative models sample their answers. Our design cannot separate these factors. Section~\ref{sec:res-luna} compares the two interfaces for one model name, and Section~\ref{sec:limitations} discusses the request design.

\paragraph{Scoring and references.} All synthetic samples are scored with the revised scoring. The gpt-5.4-mini answers and the other cells of the original audit are rescored from the stored answers with the same code. Besides the held-out human ceiling of the original audit we use four references: the real answers of the same 100 people (oracle), the in-sample copula, an out-of-sample copula fit on the 163 people outside the headline sample, and independent draws from the human marginals. The out-of-sample copula is out of sample for the 100 respondents only. Its 163 training people are part of the 263-person reference, which favours it on the distribution and correlation components. For this reason Section~\ref{sec:res-field} also scores every source against the 100 headline people alone and compares it with a second sample of 100 real people. We also read the profile regression of the forecast analysis (below) as a synthetic respondent, in the same three ways as a decision vector.

\paragraph{Uncertainty.} We use a paired bootstrap~\citep{efron1993introduction} with \vNDraws{} draws. Every draw resamples the 100 respondent IDs and the 263 humans with replacement, and the same draw is applied to every source. Resampling with replacement duplicates rows, which lowers the absolute level of every similarity component, by an amount that differs between sources. We therefore report point estimates on the original samples and use the draws only for standard errors. Intervals for paired differences are the point estimate $\pm 1.96$ bootstrap standard errors of the difference. Percentile intervals give the same conclusion for \vPctAgree{} of the \vPctN{} differences to the best generative cell and for every decision model. Paired differences are reported against two references. The first is gpt-5.4-mini, the headline model of the original audit, which was fixed as a reference before this study. The second is the best generative cell, \vBestGenName{}. This reference is the highest of \vNGen{} noisy estimates, so its point estimate is biased upward and the differences to it are biased away from zero. A source that is within noise of it is close to the top of the generative field.

\paragraph{Analysis plan.} This study was not preregistered. The benchmark, the 100 respondents, the profile fields, the PSS weights and the scoring code come from the original audit and were not changed. We added the control grids, swaps, forecast scores, reference checks, the structure analysis and the comparison of release waves after the first results and report them as exploratory. The four hypotheses were not registered either, and the tests of the third and fourth hypotheses use these later analyses. Their $p$-values are not corrected for multiple testing unless we say so.

\paragraph{Further analyses.} Response style uses raw answers for humans and models. Predictive validity fits an OLS regression of performance on attitudes and engagement on each synthetic sample and tests it on a fixed 30\% split of the humans. The null-mediation test draws \vNullDrawn{} triples of composite scores whose human indirect effect is not significant and below 0.10. We leave out the \vNullPartWhole{} triples in which a total score contains another variable of the triple, such as attitudes in total and cognitive attitudes, because they do not test mediation. The remaining \vNullN{} triples are the same for every source. A synthetic sample fabricates a path when its bootstrap interval excludes zero and the effect exceeds 0.10. This criterion does not test whether the synthetic effect differs from the small human effect. The published version of the original audit pooled all LLM cells for this test, while we run it per model at $n{=}100$. Internal structure is described by the level of the item correlations, the share of the first eigenvalue, the separation of the subscales, the subscale $\alpha$ and Tucker's $\varphi$~\citep{lorenzo2006tucker} per instrument with an item-permutation null. Weight sensitivity uses 500 Dirichlet draws around the PSS weights (concentration 30).

\paragraph{Control grids.} For the decision models we ran further grids with the same request format. One grid sends the 100 requests with an empty state, which corresponds to the condition without a profile (C0) in the original audit. Another sends the first 20 respondents again without any change, which measures run-to-run noise. Three more grids swap one field of all 100 profiles. Gender changes between woman and man, and position between managerial and non-managerial. Education follows the mapping of the original audit: university and other higher education become secondary education, and secondary and post-secondary education become university. One respondent without a gender value is left out of the gender swap. We compare the swaps with the swap grids of the original audit, which cover 37 generative models on the same respondents. These grids include Qwen 3.7 Plus. Effects are computed on the 12 composite scores in two ways: as the paired $d$ of the original audit, and as the mean shift toward the first group of the human contrast (university, managerial, woman) in units of the human SD. Both are signed by the direction of each swap. Decider v1 27B had no available endpoint when the role and gender swaps were run. Tev1 4B had by then lowered its limit to 32 questions per request, so we sent its swaps in three parts and compared them with its unchanged profiles, sent again in the same way. All control grids cost \$\vCtrlCost{}.

\paragraph{Forecast scores.} A decision model returns a forecast of how a person will answer each item. We score these forecasts against the answers that the same 100 people gave. The main score is the ranked probability score (RPS)~\citep{epstein1969scoring}. It compares the cumulative forecast with the cumulative observed answer and so respects the order of the scale points. We divide it by $K-1$ so that 5- and 7-point items can be pooled, and we also report the log score. Both scores are strictly proper~\citep{gneiting2007strictly}, so a forecaster gets the best expected score by reporting its true distribution. Before scoring, every forecast is mixed with 1\% of the uniform distribution so that the log score stays finite. We use three references. The population distribution gives every respondent the answer frequencies of the 163 humans outside the headline sample. The profile regression is a multinomial logistic regression of each item on the profile fields, fit on the same 163 people with an L2 penalty chosen by 5-fold cross-validation. The uniform distribution uses no knowledge of the survey. Skill is the relative improvement in RPS over the population distribution, with a 95\% paired bootstrap interval over respondents. We also give every respondent the model's own average vector. This tests whether the part of a vector that depends on the person improves the forecast. To separate information from calibration, we mix each vector with the population distribution and choose the weight by leave-one-respondent-out cross-fitting. Generative models return one answer per call. For them we use the test-retest panel of the original audit, in which 30 of the 100 respondents were sent 20 times to each of 31 generative models. The repeated answers form an empirical forecast for each person. We score each model on the panel members with at least 10 valid repeats, which leaves 26 to 30 people per model. To see how much of the result comes from this estimator, we also draw 20 answers from each decision vector and score them in the same way.

\section{Results}
\label{sec:results}
\subsection{Where the language models stand}
\label{sec:res-field}
\begin{table}[t]
  \centering\small
  \caption{Aligned PSS components at C3 on the same 100 respondents, scored against all 263 humans. Generative rows show the three best of 47 generative cells, the best and the worst of the 11 models released after the original audit (marked $^{\ast}$), GPT-6 Luna and the fixed reference gpt-5.4-mini; Appendix Table~\ref{tab:generative} lists all of them. The median row takes the median of each column separately. Decision models use the rounded expected score. SE is the bootstrap standard error of PSS over 200 draws that resample respondents and humans. The held-out split row is the ceiling reported in the original audit (53 humans scored against 210). The out-of-sample copula and the profile regression are fit on the 163 humans outside the headline sample; the profile regression predicts each item from the profile fields and is read out like a decision model.}
  \label{tab:leaderboard}
  \resizebox{\linewidth}{!}{%
  \begin{tabular}{lrrrrrrr}
    \toprule
    Source & Dist. & Corr. & Rel. & Med. & Demo. & PSS & SE \\
    \midrule
    \multicolumn{8}{l}{\emph{Reference points}} \\
    Same 100 humans (oracle) & 1.00 & 0.92 & 0.99 & 0.71 & 0.67 & 0.886 & 0.033 \\
    Held-out human split, original audit & 0.98 & 0.76 & 0.99 & 0.72 & 0.48 & 0.825 &  \\
    Gaussian copula, in-sample & 0.99 & 0.89 & 0.99 & 0.97 & 0.66 & 0.927 & 0.019 \\
    Gaussian copula, out-of-sample & 0.98 & 0.81 & 0.97 & 0.72 & 0.38 & 0.824 & 0.028 \\
    Independent marginals & 0.99 & 0.00 & 0.00 & 0.67 & 0.49 & 0.430 &  \\
    Profile regression, rounded & 0.64 & 0.23 & 0.78 & 0.39 & 0.36 & 0.486 &  \\
    \midrule
    \multicolumn{8}{l}{\emph{Generative LLMs}} \\
    MiniMax M3 & 0.71 & 0.76 & 0.97 & 1.00 & 0.51 & 0.813 & 0.023 \\
    MiniMax M2.7 & 0.83 & 0.81 & 0.98 & 0.76 & 0.52 & 0.809 & 0.039 \\
    Kimi K2.7 Code & 0.80 & 0.81 & 0.95 & 0.80 & 0.52 & 0.802 & 0.030 \\
    Claude Sonnet 5.5$^{\ast}$ & 0.69 & 0.70 & 0.98 & 0.73 & 0.44 & 0.735 & 0.034 \\
    GPT-6 Luna$^{\ast}$ & 0.69 & 0.57 & 0.92 & 0.46 & 0.32 & 0.621 & 0.035 \\
    gpt-5.4-mini & 0.66 & 0.60 & 0.96 & 0.93 & 0.49 & 0.743 & 0.020 \\
    GPT-6.1 Sol$^{\ast}$ & 0.69 & 0.45 & 0.85 & 0.11 & 0.33 & 0.510 & 0.030 \\
    Median of 47 generative LLMs & 0.75 & 0.73 & 0.96 & 0.46 & 0.43 & 0.695 &  \\
    \midrule
    \multicolumn{8}{l}{\emph{Decision models}} \\
    Jev 1.13 & 0.77 & 0.66 & 0.95 & 0.50 & 0.34 & 0.681 & 0.036 \\
    Microsoft-Decision-1 & 0.68 & 0.40 & 1.00 & 0.79 & 0.38 & 0.667 & 0.031 \\
    Decider v1.1 27B & 0.63 & 0.34 & 0.93 & 0.95 & 0.41 & 0.660 & 0.022 \\
    Drex v1.5 & 0.70 & 0.46 & 0.97 & 0.72 & 0.30 & 0.659 & 0.035 \\
    Decider v1 27B & 0.63 & 0.35 & 0.93 & 0.93 & 0.42 & 0.659 & 0.029 \\
    Mercury Decide & 0.78 & 0.53 & 0.95 & 0.34 & 0.48 & 0.636 & 0.035 \\
    Liquid d1 & 0.75 & 0.53 & 0.99 & 0.31 & 0.39 & 0.618 & 0.032 \\
    Clef & 0.69 & 0.54 & 0.99 & 0.09 & 0.40 & 0.563 & 0.029 \\
    Clef Flash & 0.70 & 0.62 & 0.97 & 0.00 & 0.30 & 0.554 & 0.018 \\
    Clef Omni & 0.72 & 0.50 & 0.99 & 0.22 & 0.06 & 0.551 & 0.030 \\
    GPT-6 Luna Decisions & 0.68 & 0.34 & 0.89 & 0.40 & 0.33 & 0.547 & 0.036 \\
    Kev 4B & 0.61 & 0.23 & 0.88 & 0.27 & 0.40 & 0.477 & 0.038 \\
    Tev1 4B & 0.70 & 0.24 & 0.95 & 0.00 & 0.33 & 0.460 & 0.021 \\
    Solar Decide Flash & 0.59 & 0.15 & 0.74 & 0.29 & 0.33 & 0.423 & 0.029 \\
    \bottomrule
  \end{tabular}}
\end{table}

\begin{figure}[t]
  \centering
  \includegraphics[width=\linewidth]{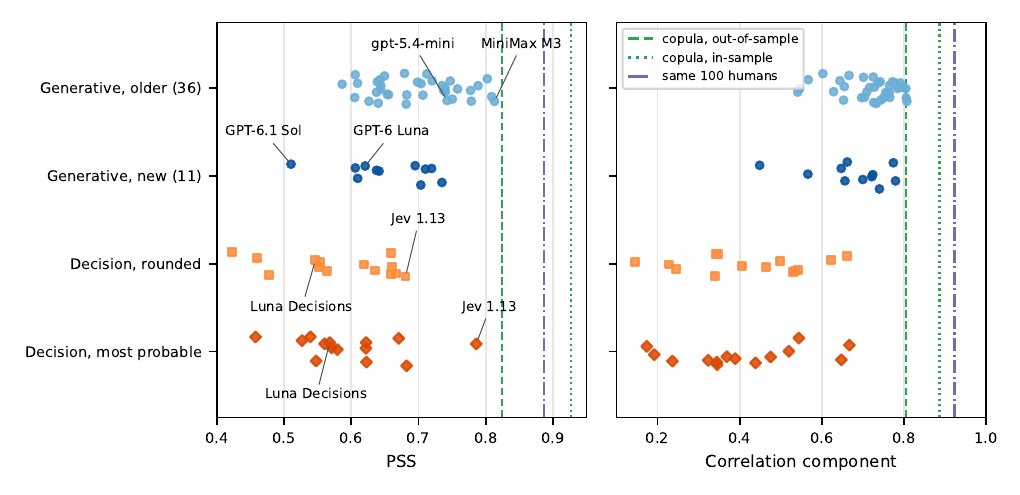}
  \caption{The best language models are level with a copula fit only on other respondents, and none reaches the in-sample copula or the oracle (the real answers of the same 100 people). Left: PSS of every language-model cell at C3. Right: the correlation component. Each point is one model, and vertical jitter only separates points. Vertical lines mark the out-of-sample copula, the in-sample copula and the oracle.}
  \label{fig:field}
\end{figure}

Figure~\ref{fig:field} and Table~\ref{tab:leaderboard} place all language models against the reference points. The oracle, the real answers of the same 100 people, reaches \vOracleTotal{}. A copula fit on other respondents reaches \vCopulaOosTotal{}, and the in-sample copula reaches \vCopulaTotal{}. The in-sample copula is above the oracle because it is fit on the 263 people it is scored against, while the 100 real respondents are a subsample with their own sampling noise.

Generative LLMs range from \vWorstNewTotal{} to \vBestGenTotal{}, with a median of \vGenMedTotal{}. The best cell is \vBestGenName{} (\vBestGenTotal), followed by \vSecondGenName{} (\vSecondGenTotal) and \vThirdGenName{} (\vThirdGenTotal). Its lead rests on the mediation component (\vBestGenMed). Without this component, with the other weights rescaled, \vBestNoMedName{} leads (\vBestNoMedTotal). The top of the field is dense. For \vNWithinBest{} generative cells the paired interval to \vBestGenName{} contains zero. The difference between the out-of-sample copula and the best cell, \vBCopulaOosTotal{} [\vBCopulaOosTotalLo, \vBCopulaOosTotalHi], is within noise. Against the oracle the interval, \vBOracleTotal{} [\vBOracleTotalLo, \vBOracleTotalHi], only just contains zero. Given the selection bias of the best cell, we do not read this as parity with real respondents. On the correlation component, the highest value of any generative cell is \vGenCorrMax{} (\vBestCorrName). The out-of-sample copula reaches a similar value (\vCopulaOosCorr), and the oracle is far higher (\vOracleCorr).

The fixed reference gpt-5.4-mini sits in the upper middle. It reaches \vGptTotal{}, with a paired difference of \vBGptTotal{} [\vBGptTotalLo, \vBGptTotalHi] to \vBestGenName{}. Its correlation component is clearly lower (\vBGptCorr{} [\vBGptCorrLo, \vBGptCorrHi]), and its total is held up by a mediation component of \vGptMed{}. The demographic component separates the models least. Independent draws from the human marginals carry no profile information and still reach 0.49 on it, above the median generative LLM.

\paragraph{A second human sample as reference.} The 263-person reference contains the 163 people on which the out-of-sample copula is fit. Appendix Table~\ref{tab:reference} also scores every source against the 100 headline people alone. On this reference the reported copula sample falls to \vRefCopOos{}, and the mean of \vNCopSeeds{} copula samples is \vRefCopSeedMean{}. A second sample of 100 real people from outside the headline sample reaches \vRefRepMean{} on average (\vRefRepMin--\vRefRepMax{} over \vNCopSeeds{} draws). The best generative cell, \vRefBestGenName{}, reaches \vRefBestGen{}, and \vRefGenAboveRep{} generative cells score above the average second human sample. Which source leads depends on the component. On the distribution component the second human sample (\vRefRepDist) and the copula (\vRefBigCopDist{} without its own sampling noise) are far ahead of every language model (at most \vRefGenDistMax). For correlation, the best generative cells (\vRefGenCorrMax) are level with the copula without sampling noise (\vRefBigCopCorr) and above the second human sample (\vRefRepCorr). The mediation and demographic components differ a lot even between two real samples of 100 people (\vRefRepMed{} and \vRefRepDemo), so a language model can score above real people there. At $n{=}100$ the top of the PSS scale therefore does not separate the best language models from real respondents.

\subsection{Decision models against the generative field}
\label{sec:res-paired}
\begin{table}[t]
  \centering\small
  \caption{Paired differences to the best generative cell, MiniMax M3 (aligned). Brackets give the point estimate $\pm 1.96$ bootstrap standard errors of the paired difference over 200 draws, clipped to $[-1, 1]$. Every draw resamples the 100 respondents and the 263 humans once and applies the same draw to both sources. R = rounded expected score, A = most probable category. An interval that contains zero means the difference is within resampling noise. The reference was chosen as the highest point estimate among 47 cells, so the differences to it are biased away from zero.}
  \label{tab:paired}
  \resizebox{\linewidth}{!}{%
  \begin{tabular}{llrrr}
    \toprule
    Source & Readout & $\Delta$ Correlation & $\Delta$ Mediation & $\Delta$ PSS \\
    \midrule
    Same 100 humans (oracle) &  & +0.16 [+0.08, +0.23] & $-$0.29 [$-$0.62, +0.04] & +0.07 [0.00, +0.15] \\
    Gaussian copula (in-sample) &  & +0.12 [+0.05, +0.20] & $-$0.03 [$-$0.24, +0.18] & +0.11 [+0.06, +0.17] \\
    Gaussian copula (out-of-sample) &  & +0.04 [$-$0.06, +0.15] & $-$0.28 [$-$0.55, $-$0.01] & +0.01 [$-$0.07, +0.09] \\
    \midrule
    MiniMax M2.7 &  & +0.04 [$-$0.01, +0.10] & $-$0.24 [$-$0.62, +0.14] & 0.00 [$-$0.09, +0.08] \\
    Kimi K2.7 Code &  & +0.04 [$-$0.01, +0.09] & $-$0.20 [$-$0.53, +0.12] & $-$0.01 [$-$0.08, +0.06] \\
    gpt-5.4-mini &  & $-$0.16 [$-$0.24, $-$0.09] & $-$0.07 [$-$0.29, +0.15] & $-$0.07 [$-$0.13, $-$0.01] \\
    GPT-6 Luna &  & $-$0.20 [$-$0.28, $-$0.12] & $-$0.54 [$-$0.86, $-$0.22] & $-$0.19 [$-$0.27, $-$0.12] \\
    \midrule
    Jev 1.13 & R & $-$0.10 [$-$0.19, $-$0.02] & $-$0.50 [$-$0.90, $-$0.10] & $-$0.13 [$-$0.21, $-$0.05] \\
    Jev 1.13 & A & $-$0.10 [$-$0.17, $-$0.03] & $-$0.03 [$-$0.33, +0.28] & $-$0.03 [$-$0.10, +0.04] \\
    Decider v1 27B & R & $-$0.42 [$-$0.49, $-$0.34] & $-$0.06 [$-$0.34, +0.21] & $-$0.15 [$-$0.22, $-$0.09] \\
    Decider v1 27B & A & $-$0.42 [$-$0.50, $-$0.34] & $-$0.55 [$-$0.91, $-$0.20] & $-$0.25 [$-$0.33, $-$0.17] \\
    Decider v1.1 27B & R & $-$0.42 [$-$0.49, $-$0.35] & $-$0.04 [$-$0.29, +0.20] & $-$0.15 [$-$0.21, $-$0.09] \\
    Decider v1.1 27B & A & $-$0.42 [$-$0.51, $-$0.33] & $-$0.45 [$-$0.87, $-$0.03] & $-$0.23 [$-$0.32, $-$0.15] \\
    Liquid d1 & R & $-$0.23 [$-$0.33, $-$0.13] & $-$0.69 [$-$1.00, $-$0.36] & $-$0.19 [$-$0.27, $-$0.12] \\
    Liquid d1 & A & $-$0.24 [$-$0.32, $-$0.17] & $-$0.48 [$-$0.84, $-$0.11] & $-$0.14 [$-$0.22, $-$0.06] \\
    Kev 4B & R & $-$0.54 [$-$0.61, $-$0.47] & $-$0.73 [$-$1.00, $-$0.34] & $-$0.34 [$-$0.42, $-$0.25] \\
    Kev 4B & A & $-$0.57 [$-$0.64, $-$0.50] & $-$0.29 [$-$0.72, +0.14] & $-$0.27 [$-$0.36, $-$0.17] \\
    Tev1 4B & R & $-$0.52 [$-$0.61, $-$0.43] & $-$1.00 [$-$1.00, $-$0.77] & $-$0.35 [$-$0.41, $-$0.29] \\
    Tev1 4B & A & $-$0.59 [$-$0.68, $-$0.50] & $-$0.64 [$-$0.92, $-$0.35] & $-$0.29 [$-$0.35, $-$0.22] \\
    Mercury Decide & R & $-$0.23 [$-$0.30, $-$0.16] & $-$0.66 [$-$1.00, $-$0.30] & $-$0.18 [$-$0.26, $-$0.10] \\
    Mercury Decide & A & $-$0.29 [$-$0.36, $-$0.22] & $-$0.69 [$-$1.00, $-$0.29] & $-$0.19 [$-$0.27, $-$0.11] \\
    Solar Decide Flash & R & $-$0.62 [$-$0.71, $-$0.53] & $-$0.70 [$-$1.00, $-$0.40] & $-$0.39 [$-$0.46, $-$0.32] \\
    Solar Decide Flash & A & $-$0.53 [$-$0.63, $-$0.43] & $-$0.13 [$-$0.56, +0.31] & $-$0.36 [$-$0.48, $-$0.23] \\
    Clef & R & $-$0.22 [$-$0.30, $-$0.14] & $-$0.91 [$-$1.00, $-$0.57] & $-$0.25 [$-$0.32, $-$0.18] \\
    Clef & A & $-$0.22 [$-$0.29, $-$0.15] & $-$1.00 [$-$1.00, $-$0.75] & $-$0.24 [$-$0.30, $-$0.19] \\
    Clef Flash & R & $-$0.14 [$-$0.20, $-$0.08] & $-$1.00 [$-$1.00, $-$0.79] & $-$0.26 [$-$0.31, $-$0.21] \\
    Clef Flash & A & $-$0.12 [$-$0.18, $-$0.06] & $-$0.76 [$-$1.00, $-$0.41] & $-$0.19 [$-$0.27, $-$0.12] \\
    Clef Omni & R & $-$0.26 [$-$0.34, $-$0.19] & $-$0.78 [$-$1.00, $-$0.46] & $-$0.26 [$-$0.33, $-$0.20] \\
    Clef Omni & A & $-$0.33 [$-$0.40, $-$0.25] & $-$0.82 [$-$1.00, $-$0.45] & $-$0.27 [$-$0.35, $-$0.19] \\
    GPT-6 Luna Decisions & R & $-$0.42 [$-$0.53, $-$0.32] & $-$0.59 [$-$0.97, $-$0.22] & $-$0.27 [$-$0.35, $-$0.19] \\
    GPT-6 Luna Decisions & A & $-$0.37 [$-$0.47, $-$0.28] & $-$0.51 [$-$0.88, $-$0.14] & $-$0.25 [$-$0.33, $-$0.16] \\
    Microsoft-Decision-1 & R & $-$0.36 [$-$0.44, $-$0.28] & $-$0.21 [$-$0.54, +0.12] & $-$0.15 [$-$0.22, $-$0.07] \\
    Microsoft-Decision-1 & A & $-$0.40 [$-$0.48, $-$0.31] & $-$0.35 [$-$0.69, $-$0.01] & $-$0.19 [$-$0.27, $-$0.11] \\
    Drex v1.5 & R & $-$0.30 [$-$0.37, $-$0.23] & $-$0.28 [$-$0.63, +0.07] & $-$0.15 [$-$0.23, $-$0.07] \\
    Drex v1.5 & A & $-$0.44 [$-$0.52, $-$0.36] & 0.00 [$-$0.20, +0.20] & $-$0.13 [$-$0.18, $-$0.08] \\
    \bottomrule
  \end{tabular}}
\end{table}

Table~\ref{tab:paired} gives the paired differences to \vBestGenName{}. Under the default rounded readout every decision model is below it, and no interval contains zero. Jev~1.13 is the best decision model with \vJevTotal{}, and its difference to the best generative cell is \vBJevTotal{} [\vBJevTotalLo, \vBJevTotalHi]. Read as the most probable category, Jev reaches \vJevArgTotal{}, and the difference of \vBJevArgTotal{} [\vBJevArgTotalLo, \vBJevArgTotalHi] is within noise. This is the best of 42 decision cells (fourteen models and three readouts), and the total rests on the mediation component (Section~\ref{sec:res-readouts}). On the correlation component Jev stays below \vBestGenName{} (\vBJevArgCorr{} [\vBJevArgCorrLo, \vBJevArgCorrHi]).

The comparison with gpt-5.4-mini does not depend on a selected reference. Under the rounded readout \vGDecBelow{} of \vGDecN{} decision models are below gpt-5.4-mini, with intervals that exclude zero. Jev read as the most probable category is \vGJevArg{} [\vGJevArgLo, \vGJevArgHi] above it, which is within noise. In the generative ranking, this readout of Jev would place \vJevArgRankGen{}th of the \vNGen{} cells (\vJevArgTotal), and its rounded readout would place \vJevRankGen{}th (\vJevTotal).

The mediation component is the least stable. Its bootstrap standard error is \vGptMedSE{} for gpt-5.4-mini and \vJevMedSE{} for Jev. Single mediation values at $n{=}100$ should not be read as model properties. Differences in PSS that come mainly from this component, such as the high rounded totals of the two Decider versions, deserve the same caution.

\subsection{GPT-6 Luna as a generator and as a decision model}
\label{sec:res-luna}
GPT-6 Luna is the only model in our set that is offered through both interfaces. We cannot verify that both endpoints serve the same weights. The two runs also differ in route (the OpenAI API against OpenRouter), in the prompt (Lithuanian text against an English JSON state with Lithuanian items) and in sampling (the provider-default temperature against a fixed endpoint). As a generator, GPT-6 Luna reaches PSS \vLunaGenTotal{} with a correlation component of \vLunaGenCorr{}. Served as a decision model, it reaches \vLunaTotal{} with the rounded readout and \vLunaArgTotal{} with the most probable category, and its correlation component falls to \vLunaCorr{}. The paired difference in correlation is \vLunaPairCorr{} [\vLunaPairCorrLo, \vLunaPairCorrHi] in favour of the generator for the rounded readout and \vLunaPairArgCorr{} [\vLunaPairArgCorrLo, \vLunaPairArgCorrHi] for the most probable category. In PSS the difference, \vLunaPair{} [\vLunaPairLo, \vLunaPairHi], only just includes zero, because the other components move less. Both interfaces give similar distribution components (\vLunaGenDist{} and \vLunaDist).

Other properties also differ. The decision answers show more agreement, with an acquiescence shift of \vLunaAcq{} human SD against \vLunaGenAcq{} for the generator. They also individualise slightly less (person share \vLunaPerson{} against \vLunaGenPerson). On the downstream regression the order reverses. A regression trained on the decision sample reaches $R^2{=}\vLunaRtwo$ on held-out humans, while the generated sample gives \vLunaGenRtwo{}. For this model name, the decision interface keeps the marginals and loses much of the correlation pattern. It also adds agreement and keeps the relation between the composite scales better.

\subsection{The readout decides which component is sacrificed}
\label{sec:res-readouts}
\begin{table}[t]
  \centering\small
  \caption{The same probability vectors read out three ways (aligned, C3, $n{=}100$): R is the expected score rounded to the nearest category, A is the most probable category, S is one category drawn from the vector per item (mean of five draws; the SD of PSS across draws is at most 0.040). Bold marks the best readout per model and metric. For reference, the median of 47 generative LLMs is 0.695 for PSS, 0.73 for correlation, 0.75 for distribution and 0.46 for mediation; the best generative cell (MiniMax M3) has PSS 0.813. The last row reads the forecasts of the profile regression, fit on the 163 humans outside the sample, in the same three ways.}
  \label{tab:readouts}
  \resizebox{\linewidth}{!}{%
  \begin{tabular}{lrrrrrrrrrrrr}
    \toprule
    & \multicolumn{3}{c}{PSS} & \multicolumn{3}{c}{Correlation} & \multicolumn{3}{c}{Distribution} & \multicolumn{3}{c}{Mediation} \\
    \cmidrule(lr){2-4}\cmidrule(lr){5-7}\cmidrule(lr){8-10}\cmidrule(lr){11-13}
    Model & R & A & S & R & A & S & R & A & S & R & A & S \\
    \midrule
    Jev 1.13 & 0.681 & \textbf{0.786} & 0.700 & 0.66 & \textbf{0.67} & 0.35 & 0.77 & 0.78 & \textbf{0.91} & 0.50 & \textbf{0.97} & 0.83 \\
    Microsoft-Decision-1 & \textbf{0.667} & 0.622 & 0.665 & \textbf{0.40} & 0.37 & 0.27 & 0.68 & 0.68 & \textbf{0.85} & 0.79 & 0.65 & \textbf{0.86} \\
    Decider v1.1 27B & \textbf{0.660} & 0.579 & 0.517 & 0.34 & \textbf{0.34} & 0.07 & 0.63 & 0.66 & \textbf{0.95} & \textbf{0.95} & 0.55 & 0.73 \\
    Drex v1.5 & 0.659 & \textbf{0.682} & 0.654 & \textbf{0.46} & 0.32 & 0.23 & 0.70 & 0.72 & \textbf{0.92} & 0.72 & \textbf{1.00} & 0.85 \\
    Decider v1 27B & \textbf{0.659} & 0.560 & 0.519 & \textbf{0.35} & 0.35 & 0.07 & 0.63 & 0.66 & \textbf{0.95} & \textbf{0.93} & 0.44 & 0.73 \\
    Mercury Decide & 0.636 & 0.623 & \textbf{0.761} & \textbf{0.53} & 0.48 & 0.46 & 0.78 & 0.79 & \textbf{0.91} & 0.34 & 0.31 & \textbf{0.87} \\
    Liquid d1 & 0.618 & \textbf{0.671} & 0.661 & \textbf{0.53} & 0.52 & 0.23 & 0.75 & 0.81 & \textbf{0.96} & 0.31 & 0.52 & \textbf{0.89} \\
    Clef & 0.563 & 0.571 & \textbf{0.615} & 0.54 & \textbf{0.54} & 0.13 & 0.69 & 0.75 & \textbf{0.95} & 0.09 & 0.00 & \textbf{0.89} \\
    Clef Flash & 0.554 & 0.622 & \textbf{0.688} & 0.62 & \textbf{0.65} & 0.35 & 0.70 & 0.75 & \textbf{0.92} & 0.00 & 0.24 & \textbf{0.83} \\
    Clef Omni & 0.551 & 0.539 & \textbf{0.654} & \textbf{0.50} & 0.44 & 0.24 & 0.72 & 0.74 & \textbf{0.92} & 0.22 & 0.18 & \textbf{0.87} \\
    GPT-6 Luna Decisions & 0.547 & 0.568 & \textbf{0.609} & 0.34 & \textbf{0.39} & 0.21 & 0.68 & 0.69 & \textbf{0.86} & 0.40 & 0.49 & \textbf{0.84} \\
    Kev 4B & 0.477 & \textbf{0.548} & 0.513 & \textbf{0.23} & 0.19 & 0.06 & 0.61 & 0.64 & \textbf{0.91} & 0.27 & 0.70 & \textbf{0.74} \\
    Tev1 4B & 0.460 & 0.527 & \textbf{0.614} & \textbf{0.24} & 0.17 & 0.16 & 0.70 & 0.70 & \textbf{0.91} & 0.00 & 0.36 & \textbf{0.90} \\
    Solar Decide Flash & 0.423 & \textbf{0.458} & 0.451 & 0.15 & \textbf{0.24} & 0.05 & 0.59 & 0.54 & \textbf{0.83} & 0.29 & \textbf{0.87} & 0.76 \\
    \midrule
    Profile regression & 0.486 & 0.616 & 0.484 & 0.23 & 0.14 & 0.03 & 0.64 & 0.74 & 0.99 & 0.39 & 0.98 & 0.64 \\
    \bottomrule
  \end{tabular}}
\end{table}

\begin{figure}[t]
  \centering
  \includegraphics[width=0.86\linewidth]{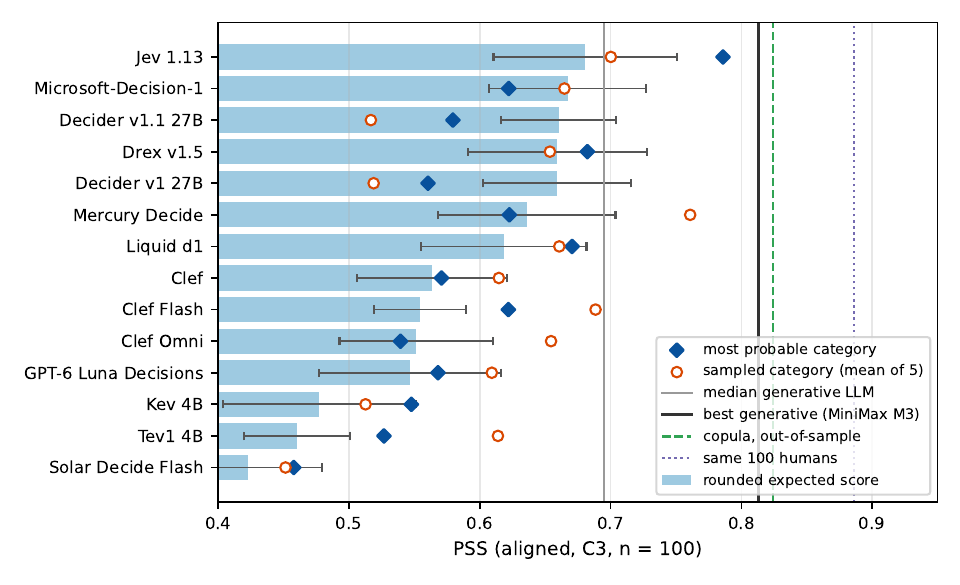}
  \caption{PSS of each decision model under three readouts of the same probability vectors. Bars show the rounded expected score with $\pm1.96$ bootstrap standard errors. Vertical lines mark the median and the best of the \vNGen{} generative LLMs, the out-of-sample copula and the real answers of the same 100 people.}
  \label{fig:readouts}
\end{figure}

\begin{figure}[t]
  \centering
  \includegraphics[width=0.78\linewidth]{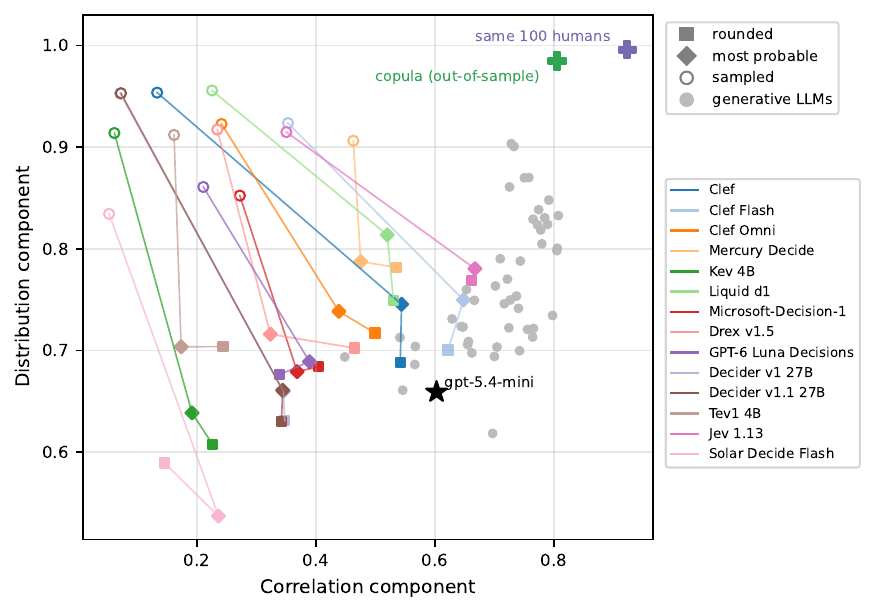}
  \caption{Sampling raises the distribution component and lowers the correlation component for every decision model. Each colour is one decision model, and lines connect its three readouts. Grey points are the \vNGen{} generative LLMs, the star is gpt-5.4-mini, and the crosses mark the out-of-sample copula and the real answers of the same 100 people.}
  \label{fig:tradeoff}
\end{figure}

Table~\ref{tab:readouts} and Figures~\ref{fig:readouts} and~\ref{fig:tradeoff} show the same vectors read three ways. The rounded expected score and the most probable category give similar correlation because both are deterministic functions of the same vector. They differ on the distribution and mediation components. Jev's gain from \vJevTotal{} to \vJevArgTotal{} comes mostly from the mediation component (\vJevMed{} to \vJevArgMed{}), which is the noisiest component.

Sampling behaves differently. It moves the distribution component of every model to 0.83--0.96, close to the copula, because each answer is now a draw from a full distribution. Sampling also lowers the correlation component of every model, by more than half for seven of them, because each item is drawn independently given the profile. Mercury Decide loses the least. It keeps a correlation component of \vMercurySampCorr{} when sampled and reaches the highest sampled PSS (\vMercurySampTotal{}), above the generative median.

Part of this loss follows from the format. When the item vectors are independent given the profile, the item correlations of a sample can only come from the profile. Tev1 4B gives a check of this independence. When its unchanged profiles are sent again in three separate requests, \vTevFormatSame\% of the rounded answers stay the same (Section~\ref{sec:limitations}). The vector for one item therefore depends little on the other items in the request. The profile regression shows how much of the correlation pattern the profile alone supports. Read out in the same three ways, it reaches correlation components of \vRegRCorr{}, \vRegACorr{} and \vRegSCorr{} (last row of Table~\ref{tab:readouts}). Under the rounded readout, \vDecAboveRegCorr{} of the \vNDecision{} decision models are above it. Decision models thus reproduce more of the correlation pattern than the profile alone supports, but less than most generative models.

The weighted total hides these trade-offs. An analyst who picks the readout after seeing PSS can choose which property of the sample to give up. This is one form of the analytic flexibility that \citet{cummins2026threat} found for silicon samples, where configurations that did well on one criterion often did poorly on another. For this reason the readout should be fixed and reported before scoring.

\subsection{What the probability vectors contain}
\label{sec:res-vectors}
\begin{table}[t]
  \centering\small
  \caption{What the probability vectors contain (aligned, C3, $n{=}100$). Entropy is normalised by the log of the number of categories, for forward and reverse-keyed items. \emph{Between} is the share of the model's answer variance (between-respondent variance of the expected score plus mean within-respondent variance of the vector) that depends on the respondent. \emph{Person} is the share of respondent-by-item variance in rounded answers that is not an item main effect. \emph{SD ratio} is the median item SD relative to all 263 humans. \emph{Unique} counts distinct 68-item answer vectors. The last two columns give the mean item Jensen-Shannon divergence to the human marginal for the rounded answers and for the average probability vector.}
  \label{tab:probabilities}
  \resizebox{\linewidth}{!}{%
  \begin{tabular}{lrrrrrrrrr}
    \toprule
    & \multicolumn{2}{c}{Entropy} & & & & & & \multicolumn{2}{c}{JS to human} \\
    \cmidrule(lr){2-3}\cmidrule(lr){9-10}
    Model & Fwd. & Rev. & Max $p$ & Between & Person & SD ratio & Unique & Rounded & Mixture \\
    \midrule
    Mercury Decide & 0.61 & 0.61 & 0.58 & 0.30 & 0.49 & 0.59 & 99 & 0.218 & 0.085 \\
    Jev 1.13 & 0.67 & 0.74 & 0.53 & 0.16 & 0.49 & 0.46 & 100 & 0.231 & 0.078 \\
    Tev1 4B & 0.81 & 0.87 & 0.40 & 0.10 & 0.40 & 0.44 & 100 & 0.296 & 0.082 \\
    Clef Flash & 0.82 & 0.79 & 0.42 & 0.12 & 0.39 & 0.46 & 94 & 0.300 & 0.068 \\
    Drex v1.5 & 0.82 & 0.84 & 0.39 & 0.12 & 0.38 & 0.45 & 99 & 0.298 & 0.076 \\
    Liquid d1 & 0.77 & 0.88 & 0.40 & 0.09 & 0.38 & 0.41 & 99 & 0.251 & 0.037 \\
    GPT-6 Luna Decisions & 0.47 & 0.66 & 0.65 & 0.13 & 0.36 & 0.32 & 99 & 0.324 & 0.130 \\
    Clef Omni & 0.75 & 0.78 & 0.43 & 0.12 & 0.32 & 0.44 & 98 & 0.283 & 0.070 \\
    Microsoft-Decision-1 & 0.72 & 0.72 & 0.47 & 0.18 & 0.30 & 0.48 & 100 & 0.316 & 0.137 \\
    Clef & 0.82 & 0.84 & 0.40 & 0.05 & 0.28 & 0.28 & 80 & 0.312 & 0.038 \\
    Kev 4B & 0.88 & 0.91 & 0.36 & 0.03 & 0.25 & 0.26 & 93 & 0.393 & 0.080 \\
    Decider v1 27B & 0.81 & 0.90 & 0.40 & 0.02 & 0.20 & 0.24 & 91 & 0.369 & 0.040 \\
    Decider v1.1 27B & 0.81 & 0.90 & 0.40 & 0.02 & 0.20 & 0.24 & 93 & 0.370 & 0.040 \\
    Solar Decide Flash & 0.75 & 0.77 & 0.47 & 0.04 & 0.15 & 0.35 & 100 & 0.411 & 0.156 \\
    \midrule
    gpt-5.4-mini (LLM) & & & & & 0.20 & 0.28 & 95 & & \\
    GPT-6 Luna (LLM) & & & & & 0.40 & & & & \\
    Median of 47 generative LLMs & & & & & 0.53 & & & & \\
    Same 100 humans & & & & & 0.81 & 1.03 & 100 & & \\
    \bottomrule
  \end{tabular}}
\end{table}

Table~\ref{tab:probabilities} shows what the probability vectors contain. First, they carry population information. Averaged over respondents, they are close to the human item marginals (Jensen-Shannon divergence \vDecJsMixMin--\vDecJsMixMax), much closer than the rounded answers (\vDecJsRoundMin--\vDecJsRoundMax). Second, only \vDecBetweenMin--\vDecBetweenMax{} of the answer variance depends on the respondent. In rounded answers, the share of variance that is not an item main effect is \vDecPersonMin--\vDecPersonMax{}, against \vHumanPerson{} for the same 100 humans and a median of \vGenMedPerson{} for the generative LLMs. Third, twelve of the fourteen models are less certain on reverse-keyed items, and for every model the reverse-keyed items keep the larger divergence after alignment (Appendix Table~\ref{tab:keying}). For Mercury Decide and Microsoft-Decision-1 the difference in entropy is below 0.01, and Clef Flash is more certain on reverse-keyed items.

A small share does not mean that the person signal is missing. The profile itself explains little of how people answer. A regression on the profile, fit on 163 real people, gives a between-respondent share of only \vRegBetween{}, and every decision model is above it (\vDecBetweenRatioMin{} to \vDecBetweenRatioMax{} times as large). The vectors also rank respondents about as well as this regression. For the decision models the mean item-level Spearman correlation between the expected score and the person's answer is \vDecRhoMin--\vDecRhoMax{}, \vRegRho{} [\vRegRhoLo, \vRegRhoHi] for the profile regression and \vGenRhoMin--\vGenRhoMax{} for the single answers of the generative models. Composites built from the expected scores correlate with the same people's real composites at a median $r$ of \vDecCompRMin--\vDecCompRMax{}, against \vRegCompR{} for the regression.

Taken together, decision models behave like a population model of item responses with a person signal on top. The signal ranks people about as well as the profile regression does, but it moves them further apart than the profile supports. Rounding turns this signal into whole scale steps, and sampling adds item-level noise that is larger than the signal. These two effects explain the readout pattern in Section~\ref{sec:res-readouts}.

\subsection{Correlation level and factor structure}
\label{sec:res-structure}
\begin{table}[t]
  \centering\small
  \caption{Internal structure of each sample (aligned answers, rounded readout for language models, the same 100 profiles). Mean $r$ is the mean inter-item correlation over all 68 items. $\lambda_1$ is the share of the largest eigenvalue of the item correlation matrix. W$-$B is the mean within-subscale minus the mean between-subscale item correlation inside an instrument, averaged over the four instruments. $\alpha$ is the median Cronbach's $\alpha$ of the 12 subscales. $\varphi$ is the median over the four instruments of Tucker's $\varphi$ against the 263 humans, with three varimax factors per instrument. \emph{Null} counts the instruments on which $\varphi$ is above an item-permutation null ($K{=}500$, $p{<}0.05$). Items without variance in a sample are left out of its $\varphi$. $\lambda_1$(E) is the same eigenvalue share for the expected scores of a decision model. The oracle is part of the 263-person reference, which favours it.}
  \label{tab:structure}
  \setlength{\tabcolsep}{4pt}
  \begin{tabular}{lrrrrrrr}
    \toprule
    Source & Mean $r$ & $\lambda_1$ & W$-$B & $\alpha$ & $\varphi$ & Null & $\lambda_1$(E) \\
    \midrule
    Humans ($n{=}263$, reference) & 0.22 & 0.28 & 0.16 & 0.81 & 1.00 & 4 &  \\
    Same 100 humans (oracle) & 0.25 & 0.33 & 0.15 & 0.81 & 0.97 & 4 &  \\
    Copula, in-sample & 0.21 & 0.27 & 0.12 & 0.79 & 0.96 & 4 &  \\
    Copula, out-of-sample & 0.15 & 0.20 & 0.14 & 0.79 & 0.91 & 3 &  \\
    \midrule
    MiniMax M3 (best LLM) & 0.37 & 0.44 & 0.16 & 0.89 & 0.87 & 3 &  \\
    gpt-5.4-mini & 0.23 & 0.29 & 0.19 & 0.72 & 0.69 & 2 &  \\
    GPT-6 Luna (LLM) & 0.24 & 0.32 & 0.16 & 0.77 & 0.78 & 3 &  \\
    Median of 47 generative LLMs & 0.36 & 0.49 & 0.17 & 0.88 & 0.82 & 3 &  \\
    \midrule
    Jev 1.13 & 0.41 & 0.47 & 0.13 & 0.88 & 0.78 & 3 & 0.68 \\
    Decider v1 27B & 0.23 & 0.25 & 0.12 & 0.58 & 0.58 & 0 & 0.60 \\
    Decider v1.1 27B & 0.23 & 0.24 & 0.11 & 0.60 & 0.60 & 1 & 0.60 \\
    Liquid d1 & 0.31 & 0.38 & 0.10 & 0.79 & 0.72 & 1 & 0.65 \\
    Kev 4B & 0.21 & 0.26 & 0.07 & 0.53 & 0.66 & 1 & 0.58 \\
    Tev1 4B & 0.32 & 0.42 & 0.02 & 0.78 & 0.58 & 1 & 0.71 \\
    Mercury Decide & 0.30 & 0.48 & 0.24 & 0.88 & 0.72 & 1 & 0.62 \\
    Solar Decide Flash & 0.13 & 0.32 & 0.02 & 0.68 & 0.39 & 0 & 0.65 \\
    Clef & 0.32 & 0.36 & 0.11 & 0.79 & 0.72 & 1 & 0.76 \\
    Clef Flash & 0.42 & 0.49 & 0.14 & 0.85 & 0.76 & 2 & 0.83 \\
    Clef Omni & 0.44 & 0.47 & 0.10 & 0.81 & 0.79 & 0 & 0.85 \\
    GPT-6 Luna Decisions & 0.23 & 0.29 & 0.04 & 0.66 & 0.68 & 3 & 0.61 \\
    Microsoft-Decision-1 & 0.25 & 0.37 & 0.16 & 0.75 & 0.64 & 2 & 0.52 \\
    Drex v1.5 & 0.24 & 0.36 & 0.14 & 0.75 & 0.63 & 3 & 0.56 \\
    \bottomrule
  \end{tabular}
\end{table}

The correlation component compares the pattern of two correlation matrices. Table~\ref{tab:structure} adds their level and the factor structure. Generative LLMs make the items correlate too strongly. Their median mean inter-item correlation is \vStrGenRMed{}, against \vStrHumanR{} for the 263 humans and \vStrOracleR{} for the same 100 people, and \vStrGenRAbove{} of the \vNGen{} cells are above the value of these 100 people. The first eigenvalue carries a median share of \vStrGenEigMed{} of the variance, against \vStrHumanEig{} and \vStrOracleEig{}, and the median subscale $\alpha$ is \vStrGenAlphaMed{} against \vStrHumanAlpha{}. The original audit described the same over-coherence.

Most decision models stay closer to the human level of correlation (median \vStrDecRMed{}, range \vStrDecRMin--\vStrDecRMax). They separate the subscales less well, however. In humans the mean correlation within a subscale exceeds the mean correlation between subscales by \vStrHumanWB{}. For \vStrDecWBBelow{} of the \vNDecision{} decision models this margin is smaller, down to \vStrDecWBMin{}. Tucker's $\varphi$ gives the same picture. Its median over the four instruments is \vStrDecPhiMin--\vStrDecPhiMax{} for the decision models, \vStrGenPhiMed{} for the median generative LLM and \vStrOraclePhi{} for the same 100 people. Every decision model is below 0.85, the usual threshold for a fair similarity of factor structures~\citep{lorenzo2006tucker}.

The expected scores show why. In them the first eigenvalue carries \vStrDecEigExpMin--\vStrDecEigExpMax{} of the variance, so the person signal of the vectors moves nearly all items in one direction. The profile sets a general level for each respondent, and the vectors add little that is specific to a subscale. This gives a more precise reading of the second hypothesis. In most decision samples the item correlations are close to the human level, but they follow one general direction instead of the subscale structure.

\subsection{The person signal is stable and comes from the profile}
\label{sec:res-controls}
\begin{table}[t]
  \centering\small
  \caption{Control grids for the decision models. \emph{Same} is the percentage of rounded answers that are identical when 20 respondents are sent again. \emph{Repl.} is the share of between-respondent variance of the expected score that is not run-to-run noise. \emph{Shift} is the mean absolute difference between the C3 expected score and the no-profile (C0) expected score on a 0--1 scale. PSS uses the sampled readout, which is the only readout that gives varying answers at C0, averaged over the same five draws as in Table~\ref{tab:readouts}. The JS columns give the mean item Jensen-Shannon divergence between the human marginal and the average probability vector. The generative row uses the C0 and C3 cells of the original audit. GPT-6 Luna Decisions returned HTTP 502 for every request without a profile.}
  \label{tab:controls}
  \begin{tabular}{lrrrrrrr}
    \toprule
    & \multicolumn{2}{c}{Retest} & \multicolumn{3}{c}{Profile} & \multicolumn{2}{c}{JS to human} \\
    \cmidrule(lr){2-3}\cmidrule(lr){4-6}\cmidrule(lr){7-8}
    Model & Same & Repl. & Shift & PSS C0 & PSS C3 & C0 & C3 \\
    \midrule
    Jev 1.13 & 95 & 0.99 & 0.14 & 0.40 & 0.70 & 0.154 & 0.078 \\
    Microsoft-Decision-1 & 95 & 0.98 & 0.16 & 0.39 & 0.66 & 0.206 & 0.137 \\
    Decider v1.1 27B & 100 & 1.00 & 0.04 & 0.43 & 0.52 & 0.046 & 0.040 \\
    Drex v1.5 & 99 & 1.00 & 0.18 & 0.40 & 0.65 & 0.156 & 0.076 \\
    Decider v1 27B & 100 & 1.00 & 0.04 & 0.43 & 0.52 & 0.046 & 0.040 \\
    Mercury Decide & 99 & 1.00 & 0.13 & 0.43 & 0.76 & 0.081 & 0.085 \\
    Liquid d1 & 92 & 0.94 & 0.08 & 0.42 & 0.66 & 0.055 & 0.037 \\
    Clef & 100 & 1.00 & 0.06 & 0.43 & 0.61 & 0.047 & 0.038 \\
    Clef Flash & 100 & 1.00 & 0.18 & 0.40 & 0.69 & 0.199 & 0.068 \\
    Clef Omni & 100 & 1.00 & 0.10 & 0.43 & 0.65 & 0.075 & 0.070 \\
    GPT-6 Luna Decisions & 100 & 1.00 &  &  & 0.61 &  & 0.130 \\
    Kev 4B & 100 & 1.00 & 0.09 & 0.42 & 0.51 & 0.091 & 0.080 \\
    Tev1 4B & 98 & 1.00 & 0.14 & 0.40 & 0.61 & 0.166 & 0.082 \\
    Solar Decide Flash & 89 & 0.82 & 0.14 & 0.39 & 0.45 & 0.251 & 0.156 \\
    \midrule
    Median of 37 generative LLMs & & & & \multicolumn{2}{c}{gain +0.16} & & \\
    \bottomrule
  \end{tabular}
\end{table}

Table~\ref{tab:controls} reports the three control grids. The repeated requests show that most decision models are deterministic. Both Decider versions, the three Clef models and GPT-6 Luna Decisions returned the same vectors, and Kev 4B, Tev1 4B, Mercury Decide and Drex v1.5 changed them only in the second decimal. Four models vary between runs. Of their rounded answers, \vJevRetestSame\% repeat for Jev~1.13, \vDOneRetestSame\% for Liquid d1, \vMsRetestSame\% for Microsoft-Decision-1 and \vSolarRetestSame\% for Solar Decide Flash. Even for these four models, run-to-run noise is small next to the differences between respondents. At least \vReplMin{} of the between-respondent variance of the expected score repeats, and the composite scores of the two runs correlate at \vRetestCompRMin{} or more. The person signal of Section~\ref{sec:res-vectors} is a stable property of the answers.

Without a profile, the deterministic models give every respondent the same 68 answers. Their rounded and most probable readouts then have no item variance, and the rounded PSS falls to \vCzeroRoundMin--\vCzeroRoundMax. Jev~1.13 returned \vJevCzeroUnique{} different rounded vectors to 100 identical requests, but this variation equals only \vJevCzeroNoise{} of its between-respondent variance at C3. Solar Decide Flash is noisier. It returned \vSolarCzeroUnique{} different rounded vectors, and their variation equals \vSolarCzeroNoise{} of its between-respondent variance at C3. The sampled readout varies at C0 as well. Its PSS rises from \vCzeroSampMin--\vCzeroSampMax{} at C0 to the C3 values, a median gain of \vGainMed{} (range \vGainMin{} to \vGainMax), so decision models do read the profile. The generative models of the original audit gain a median of \vGenGainMed{} between the same two conditions, but the two gains are not directly comparable. Sampling at C0 removes all item correlation from a decision model, while generative models keep some at C0. On average the profile moves the expected score by \vShiftMin--\vShiftMax{} of the scale range. Decision and generative models differ in the size of this shift relative to the spread inside each vector.

The profile also carries part of the population information. For \vJsWorse{} of the \vNCzero{} models with a C0 grid, the C0 vector is further from the human marginals than the vector averaged over the 100 C3 profiles. The close match of the averaged vectors in Table~\ref{tab:probabilities} depends on averaging over real profiles as well as on the item text. For both Decider versions, Clef, Clef Omni and Kev 4B the gap is small, and for Mercury Decide the C0 vector is slightly closer.

\subsection{Counterfactual swaps on three axes}
\label{sec:res-swaps}
\begin{table}[t]
  \centering\small
  \caption{Counterfactual swaps of one profile field, averaged over the 12 composites. $|$Shift$|$ is the mean absolute shift toward the first group of the human contrast (university, managerial, woman) in human SD units. $|d|$ is the mean absolute paired $d$ of the original audit, with each difference signed by the direction of the swap. Decision models use the rounded readout. Generative rows use the swap grids of the original audit. The human row gives the mean $|d|$ of the observed contrast between the two groups among all 263 humans, which is on the same scale as $|$Shift$|$. The education contrast rests on 15 respondents with secondary education, and the gender contrast is null. Section~\ref{sec:res-swaps} gives bootstrap intervals for the medians and the human contrasts. Decider v1 27B had no available endpoint when the role and gender swaps were run.}
  \label{tab:swaps}
  \begin{tabular}{lrrrrrr}
    \toprule
    & \multicolumn{2}{c}{Education} & \multicolumn{2}{c}{Role} & \multicolumn{2}{c}{Gender} \\
    \cmidrule(lr){2-3}\cmidrule(lr){4-5}\cmidrule(lr){6-7}
    Model & $|$Shift$|$ & $|d|$ & $|$Shift$|$ & $|d|$ & $|$Shift$|$ & $|d|$ \\
    \midrule
    Jev 1.13 & 0.33 & 1.11 & 0.22 & 0.92 & 0.01 & 0.11 \\
    Microsoft-Decision-1 & 0.30 & 1.36 & 0.33 & 1.54 & 0.03 & 0.17 \\
    Decider v1.1 27B & 0.15 & 1.00 & 0.18 & 1.43 & 0.03 & 0.34 \\
    Drex v1.5 & 0.37 & 1.39 & 0.28 & 1.28 & 0.08 & 0.54 \\
    Decider v1 27B & 0.15 & 1.01 &  &  &  &  \\
    Mercury Decide & 0.21 & 0.92 & 0.19 & 0.50 & 0.16 & 0.46 \\
    Liquid d1 & 0.36 & 1.29 & 0.32 & 1.17 & 0.04 & 0.19 \\
    Clef & 0.15 & 0.89 & 0.14 & 0.83 & 0.02 & 0.31 \\
    Clef Flash & 0.23 & 1.19 & 0.62 & 2.78 & 0.04 & 0.40 \\
    Clef Omni & 0.15 & 0.83 & 0.49 & 2.33 & 0.03 & 0.24 \\
    GPT-6 Luna Decisions & 0.12 & 0.62 & 0.16 & 0.85 & 0.02 & 0.16 \\
    Kev 4B & 0.25 & 1.10 & 0.17 & 0.88 & 0.08 & 0.58 \\
    Tev1 4B & 0.31 & 1.44 & 0.31 & 1.70 & 0.10 & 0.72 \\
    Solar Decide Flash & 0.20 & 0.94 & 0.12 & 0.67 & 0.03 & 0.17 \\
    \midrule
    Median of decision models & 0.22 & 1.06 & 0.22 & 1.17 & 0.03 & 0.31 \\
    Median of 37 generative LLMs & 0.32 & 0.68 & 0.47 & 1.05 & 0.06 & 0.17 \\
    Humans, observed contrast & 0.44 &  & 0.46 &  & 0.11 &  \\
    \bottomrule
  \end{tabular}
\end{table}

Table~\ref{tab:swaps} reports the swaps. For role and education the decision models move the composites in the direction of the human contrast in \vRoleSignDec\% and \vEduSignDec\% of model-composite pairs (\vRoleSignGen\% and \vEduSignGen\% for the generative models). In humans and in models, managers and university graduates report more positive attitudes to change, more engagement and higher contextual performance. The size of the reaction differs between the two classes. Measured in human SD units, the decision models shift by \vRoleShiftMin--\vRoleShiftMax{} after a role swap (median \vRoleShiftMed, 95\% bootstrap interval \vRoleShiftMedLo--\vRoleShiftMedHi) and by \vEduShiftMin--\vEduShiftMax{} after an education swap (median \vEduShiftMed). The \vNGenEdu{} generative models shift by a median of \vGenRoleShiftMed{} after a role swap and \vGenEduShiftMed{} after an education swap. The observed human difference is \vHumanRoleD{} [\vHumanRoleDLo, \vHumanRoleDHi] for role, where 7 of 12 contrasts survive a Holm correction. For education it is \vHumanEduD{} [\vHumanEduDLo, \vHumanEduDHi], but it rests on 15 people with secondary education, and none of its contrasts survives the correction. Generative models reproduce the role difference at about its observed size. The interval for the median decision model lies below the interval of the human role contrast. Of the \vNRoleDec{} decision models, \vRoleBelowHalf{} reach less than half of it, and \vRoleBelowGen{} shift less than the weakest generative model. Role moves the decision models more than education in only \vRoleOverEdu{} of \vNRoleEdu{} models, while the original audit found role to be the stronger axis in 34 of 37 generative models.

The gender swap moves the composites by \vGenderShiftMin--\vGenderShiftMax{} human SD in the decision models (median \vGenderShiftMed) and by a median of \vGenGenderShiftMed{} in the generative models. In humans the gender contrast is null, and the direction of the model shifts agrees with the sign of the null human estimates in about half of the composites (\vGenderSignDec\% and \vGenderSignGen\%). Neither class adds a gender effect of practical size.

The paired $d$ of the original audit, with each difference signed by the direction of the swap, can reverse the order of the two classes. For education it is \vEduDMed{} for the median decision model against \vGenEduDMed{} for the median generative model, and for gender it is \vGenderDMed{} against \vGenGenderDMed{}. The paired $d$ divides the mean shift by the spread of the shifts across respondents. A nearly deterministic model shifts every respondent in almost the same way, so this spread is small and $d$ is large. The paired $d$ thus measures how uniform a reaction is rather than how large it is. It should not be compared across models with different levels of answer noise.

\subsection{The vectors as forecasts of each person's answers}
\label{sec:res-forecast}
\begin{table}[t]
  \centering\small
  \caption{Probability vectors scored as forecasts of the answers of the same 100 people. Skill is $100\,(1 - \mathrm{RPS}/\mathrm{RPS}_{\text{pop}})$, where RPS is the ranked probability score averaged over items and respondents and $\mathrm{RPS}_{\text{pop}} = 0.128$ is the score of the population distribution of the 163 humans outside the sample. Negative values are worse than this distribution. Brackets give 95\% paired bootstrap intervals over respondents. \emph{Shrunk} mixes each vector with the population distribution, with the weight chosen by leave-one-respondent-out cross-fitting. \emph{Own average} gives every respondent the model's mean vector over all respondents. \emph{C0} uses the vectors returned without a profile, and \emph{Rounded} scores the rounded expected score as a certain forecast. $\rho$ is the mean item-level Spearman correlation between the expected score and the person's answer. Entropy is the mean entropy of the forecast in nats. The profile regression is a multinomial logistic regression of each item on the profile fields, fit on the same 163 humans.}
  \label{tab:forecast}
  \resizebox{\linewidth}{!}{%
  \begin{tabular}{lrrrrrrr}
    \toprule
    Forecaster & Skill (\%) & Shrunk & Own avg. & C0 & Rounded & $\rho$ & Entropy \\
    \midrule
    Clef & $-$6 [$-$10, $-$2] & 0.6 & $-$7 & $-$9 & $-$48 & 0.15 & 1.40 \\
    Liquid d1 & $-$8 [$-$12, $-$3] & 0.9 & $-$9 & $-$16 & $-$51 & 0.16 & 1.35 \\
    Decider v1.1 27B & $-$8 [$-$12, $-$4] & 0.1 & $-$10 & $-$10 & $-$55 & 0.14 & 1.40 \\
    Decider v1 27B & $-$8 [$-$12, $-$4] & 0.1 & $-$10 & $-$10 & $-$55 & 0.14 & 1.41 \\
    Clef Omni & $-$16 [$-$23, $-$10] & 0.8 & $-$17 & $-$16 & $-$58 & 0.18 & 1.29 \\
    Clef Flash & $-$18 [$-$26, $-$11] & 0.7 & $-$18 & $-$75 & $-$65 & 0.19 & 1.39 \\
    Jev 1.13 & $-$18 [$-$25, $-$12] & 0.7 & $-$15 & $-$40 & $-$57 & 0.17 & 1.17 \\
    GPT-6 Luna Decisions & $-$22 [$-$26, $-$17] & 0.0 & $-$21 &  & $-$55 & 0.16 & 0.88 \\
    Kev 4B & $-$22 [$-$29, $-$15] & $-$0.4 & $-$23 & $-$29 & $-$73 & 0.11 & 1.50 \\
    Tev1 4B & $-$23 [$-$30, $-$17] & 0.1 & $-$25 & $-$51 & $-$70 & 0.18 & 1.40 \\
    Drex v1.5 & $-$24 [$-$32, $-$16] & $-$0.1 & $-$22 & $-$49 & $-$72 & 0.13 & 1.41 \\
    Mercury Decide & $-$32 [$-$40, $-$25] & 0.4 & $-$21 & $-$24 & $-$75 & 0.13 & 1.05 \\
    Solar Decide Flash & $-$49 [$-$58, $-$41] & $-$0.5 & $-$49 & $-$69 & $-$104 & 0.11 & 1.29 \\
    Microsoft-Decision-1 & $-$55 [$-$67, $-$43] & 0.0 & $-$51 & $-$81 & $-$105 & 0.10 & 1.23 \\
    \midrule
    Profile regression & 1.0 [0.1, 1.8] & & & & & 0.15 & 1.35 \\
    Population distribution & 0 & & & & &  & 1.36 \\
    Uniform & $-$42 [$-$53, $-$32] & & & & &  & 1.69 \\
    \bottomrule
  \end{tabular}}
\end{table}

Table~\ref{tab:forecast} scores the probability vectors against the answers of the same people. The profile itself carries little information about a person's answers. A regression on the profile fields beats the population distribution of other respondents by only \vFcRegGain\% [\vFcRegLo, \vFcRegHi]. No decision model reaches the population distribution. Their vectors are \vFcDecWorseMin--\vFcDecWorseMax\% worse (median \vFcDecWorseMed\%), and the interval lies below zero for all \vFcNDec{} models. They are still much better than the uniform distribution, which is \vFcUniWorse\% worse. Rounding a vector to one answer loses most of this advantage. The rounded answers are \vFcRoundWorseMin--\vFcRoundWorseMax\% worse than the population distribution.

Most of this loss comes from the average vector. Given to every respondent, the model's own average vector is \vFcPoolWorseMin--\vFcPoolWorseMax\% worse than the population distribution (column Own avg.), about as bad as the individual vectors. For some models the person-specific part adds a further loss, because the vectors spread respondents further apart than the profile supports (Section~\ref{sec:res-vectors}). The log score gives the same verdict. All \vFcDecLogNeg{} decision models are worse than the population distribution under it, by \vFcDecLogWorseMin--\vFcDecLogWorseMax\%, and GPT-6 Luna Decisions, Solar Decide Flash, Microsoft-Decision-1 and Mercury Decide lose most. All of them except Solar Decide Flash are among the four most confident models. We mixed each vector with the population distribution. Cross-fitting chose a weight of \vFcShrinkLamMin--\vFcShrinkLamMax{} on the vector. This brings the skill to \vFcShrinkMin{} to \vFcShrinkMax\%, close to the profile regression. No shrunk vector has a skill interval that lies above zero.

The profile improves the forecast for \vFcCthreeBetter{} of the \vFcNCzero{} models with a C0 grid. The individual vectors beat the model's own average vector in \vFcPoolBetter{} of \vFcNDec{} models, by a median of \vFcPoolDiffMed{} percentage points. Most of the benefit of the profile thus comes from moving the average vector closer to the population, which agrees with the C0 result of Section~\ref{sec:res-controls}. Differences between the vectors of different respondents do carry information about the person, as the rank correlations show. Without recalibration this information adds little to the forecast, because the vectors spread respondents too far apart.

\begin{figure}[t]
  \centering
  \includegraphics[width=0.86\linewidth]{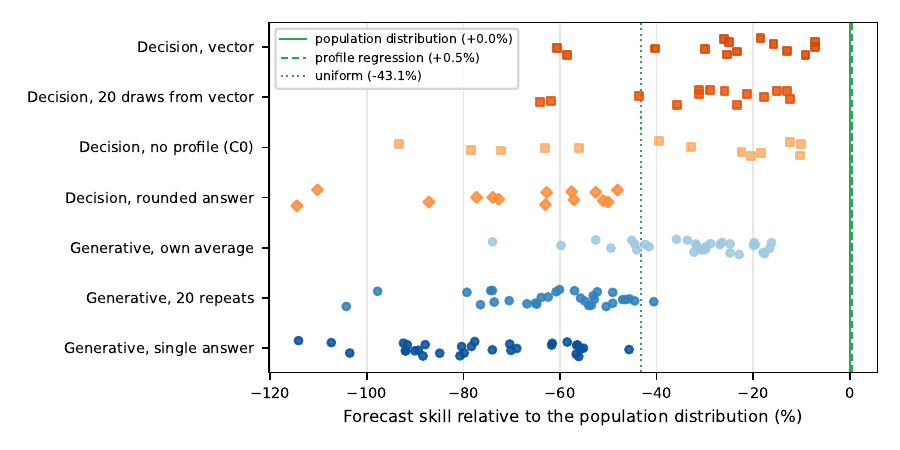}
  \caption{No language model forecasts a person's answers better than the answer frequencies of other respondents. Forecast skill on the 26 to 30 respondents of the test-retest panel, relative to the population distribution of the 163 humans outside the headline sample. Each point is one model. Vertical lines mark the population distribution (zero), the profile regression and the uniform distribution. The second row draws 20 answers from each decision vector and scores them like the generative repeats.}
  \label{fig:forecast}
\end{figure}

Figure~\ref{fig:forecast} adds the generative models on the panel. Their repeat distributions are \vFcGenWorseMin--\vFcGenWorseMax\% worse than the population distribution (median \vFcGenWorseMed\%). Only a small part of this gap comes from the estimator. Twenty draws from a decision vector, scored in the same way, are \vFcDrawWorseMin--\vFcDrawWorseMax\% worse, a median loss of \vFcDrawCostMed{} points against the full vector. Most of the gap comes from the repeat distributions themselves, which are too narrow. Their median entropy is \vFcGenEntMed{} nats, against \vFcPopEnt{} for the population distribution and \vFcDecEntMed{} for the decision vectors. A single answer, which is what a generative survey run uses, is worse still (median \vFcGenDrawWorseMed\%). In all \vFcGenPoolBetter{} generative models, the pooled answers of the panel forecast each person better than that person's own repeats, by a median of \vFcGenPoolDiffMed{} points. Shrinking each person's repeats toward the model's pooled answers, with the weight again chosen by cross-fitting, gives \vFcGenShrinkMin{} to \vFcGenShrinkMax\% (median \vFcGenShrinkMed\%). Decision vectors are the better forecasts before recalibration. Once both classes are recalibrated, neither does much better than the profile regression, which uses the same fields.

\subsection{Newer generative models do not score higher}
\label{sec:res-newer}
\begin{table}[t]
  \centering\small
  \caption{Aligned PSS of successive releases in each product line with a release after the original audit, oldest first. $\Delta$ compares the two releases in bold: the newest release with the newest release of the same line in the original audit. Brackets give the 95\% paired bootstrap interval. In 8 of 10 lines the newest release scores lower, 1 is level and 1 gains. In 5 lines the interval lies below zero. \emph{No M} is $\Delta$ for PSS without the mediation component, with the other weights rescaled; it is negative in 9 lines. The DeepSeek, Moonshot Kimi and OpenAI small lines change tier as well as version. MiniMax has no release after the original audit, and Qwen 3.7 Plus has too few complete respondents to be scored.}
  \label{tab:successors}
  \begin{tabular}{@{}lp{0.43\linewidth}rrr@{}}
    \toprule
    Line & Releases (PSS) & $\Delta$ & 95\% interval & No M \\
    \midrule
    Anthropic Opus & {Claude~Opus~4.7~0.79} $\rightarrow$ \textbf{Claude~Opus~4.8~0.74} $\rightarrow$ \textbf{Claude~Opus~5.5~0.61} & $-$0.13 & [$-$0.20, $-$0.07] & $-$0.02 \\
    Anthropic Sonnet & {Claude~Sonnet~4.6~0.59} $\rightarrow$ \textbf{Claude~Sonnet~5~0.74} $\rightarrow$ \textbf{Claude~Sonnet~5.5~0.73} & 0.00 & [$-$0.08, +0.07] & 0.00 \\
    OpenAI flagship & {GPT-5.4~0.61} $\rightarrow$ \textbf{GPT-5.5~0.61} $\rightarrow$ \textbf{GPT-6.1~Sol~0.51} & $-$0.10 & [$-$0.17, $-$0.03] & $-$0.10 \\
    OpenAI small & \textbf{gpt-5.4-mini~0.74} $\rightarrow$ \textbf{GPT-6~Luna~0.62} & $-$0.12 & [$-$0.19, $-$0.05] & $-$0.03 \\
    Google Flash & {Gemini~3~Flash~0.70} $\rightarrow$ \textbf{Gemini~3.5~Flash~0.74} $\rightarrow$ {Gemini~3.6~Flash~0.64} $\rightarrow$ \textbf{Gemini~3.8~Flash~0.71} & $-$0.02 & [$-$0.10, +0.05] & $-$0.01 \\
    xAI Grok & \textbf{Grok~4.3~0.71} $\rightarrow$ \textbf{Grok~4.7~0.61} & $-$0.10 & [$-$0.19, 0.00] & $-$0.04 \\
    DeepSeek & {DeepSeek~3.2~0.66} $\rightarrow$ \textbf{DeepSeek~V4~Pro~0.75} $\rightarrow$ \textbf{DeepSeek~V4.1~Flash~0.70} & $-$0.05 & [$-$0.15, +0.06] & $-$0.06 \\
    Moonshot Kimi & {Kimi~K2.6~0.75} $\rightarrow$ \textbf{Kimi~K2.7~Code~0.80} $\rightarrow$ \textbf{Kimi~K3~0.64} & $-$0.16 & [$-$0.23, $-$0.09] & $-$0.07 \\
    Alibaba Qwen & \textbf{Qwen3.5~397B~0.78} $\rightarrow$ \textbf{Qwen~3.8~Max~0.72} & $-$0.06 & [$-$0.15, +0.03] & $-$0.03 \\
    Zhipu GLM & {GLM~5.1~0.71} $\rightarrow$ \textbf{GLM~5.2~0.64} $\rightarrow$ \textbf{GLM~5.3~0.70} & +0.06 & [0.00, +0.12] & $-$0.04 \\
    \bottomrule
  \end{tabular}
\end{table}

The data do not support the fourth hypothesis. The \vNGenNew{} generative models released after the original audit have a median PSS of \vGenNewMedTotal{}, against \vGenFirstMedTotal{} for the \vNGenFirst{} older cells (Mann-Whitney $p{=}\vGenNewP$). Without the mediation component the medians are \vGenNewMedNoMed{} and \vGenFirstMedNoMed{} ($p{=}\vGenNewPNoMed$). The best new model, \vBestNewName{} (\vBestNewTotal), ranks \vBestNewRank{}th of \vNGen{}. \vWorstNewName{} (\vWorstNewTotal) is last of all \vNGen{} cells, and GPT-6 Luna (\vLunaGenTotal) ranks \vLunaGenRank{}th.

Table~\ref{tab:successors} compares releases within one product line, which holds the vendor and the training lineage roughly fixed. For each line we compare the newest release with the newest release of the same line in the original audit. In \vSuccDown{} of \vSuccN{} lines the newest release scores lower, \vSuccLevel{} is level and \vSuccUp{} gains (sign test $p{=}\vSuccSignP$). The median change is \vSuccMedDelta{}. For \vSuccBelow{} lines the paired interval lies below zero, and for none it lies above zero. Three lines also change size tier (Table~\ref{tab:successors}). Without them, the newest release is lower in \vSuccSameTierDown{} of the \vSuccSameTierMove{} lines that change (sign test $p{=}\vSuccSameTierP$). Claude Opus 5.5 (\vOpusFiveTotal) is below Claude Opus 4.8 (\vOpusFourEightTotal), which was already below Claude Opus 4.7 (\vOpusFourSevenTotal). Kimi K3 (\vKimiThreeTotal) is below Kimi K2.7 Code (\vKimiCodeTotal). Without the mediation component the newest release is lower in \vSuccDownNoMed{} of \vSuccN{} lines, so the direction of the result does not depend on the noisiest component. The size of single drops, such as those in the Claude Opus and Kimi lines, comes mostly from the mediation component.

One decision model also has a successor in our set. Its provider describes Decider v1.1 27B as the successor of Decider v1 27B. The two versions return nearly the same vectors, and 98\% of their rounded answers agree. Under the rounded readout their PSS is \vDeciderOldTotal{} and \vDeciderNewTotal{}, a paired difference of \vDDeciderTotal{} [\vDDeciderTotalLo, \vDDeciderTotalHi]. The correlation component changes by \vDDeciderCorr{} [\vDDeciderCorrLo, \vDDeciderCorrHi]. This release therefore changed the answers very little.

The new models differ from the old ones mainly in response style. Their median item SD relative to humans is \vWaveNewSD{} against \vWaveFirstSD{} ($p{=}\vWavePSD$). The share of extreme answers is also lower (\vWaveNewExtreme{} against \vWaveFirstExtreme, $p{=}\vWavePExtreme$), but this difference is within noise. New models may individualise less (person share \vWaveNewPerson{} against \vWaveFirstPerson, $p{=}\vWavePPerson$), and their demographic component falls as well (\vWaveNewDemo{} against \vWaveFirstDemo, $p{=}\vWavePDemo$). The differences in the correlation (\vWaveNewCorr{} against \vWaveFirstCorr) and distribution components (\vWaveNewDist{} against \vWaveFirstDist) point the same way but are smaller than the noise. On predictive validity the new models do slightly better (median $R^2$ \vWaveNewRtwo{} against \vWaveFirstRtwo, $p{=}\vWavePRtwo$). These are exploratory comparisons of two small groups. None of the \vWaveNTests{} comparisons of release waves in this section survives a Holm correction (smallest adjusted $p{=}\vWaveHolmMin$). The pattern suggests that newer models answer in a narrower and more uniform way, which costs fidelity on most components.

\subsection{Response style and downstream use}
\label{sec:res-downstream}
\begin{table}[t]
  \centering\small
  \caption{Response style on a common raw orientation (answers to the original wording for humans and models). ARS is the balanced acquiescence index: the mean of the forward-item and reverse-item means on the 0--1 scale, minus 0.5, over the two instruments with reverse-keyed items. $\Delta$ ARS is the difference to humans in units of the human ARS standard deviation. \emph{Fwd} and \emph{Rev} are the differences of the forward-item mean and of the reverse-item mean in the same units, so that $\Delta$ ARS is their average. The keying gap is the ``acquiescence'' index of the original audit (forward mean minus reverse mean), which grows with content consistency. Extreme and midpoint rates are shares of answers at the scale ends and the centre. SD ratio is the median item standard deviation relative to humans.}
  \label{tab:style}
  \resizebox{\linewidth}{!}{%
  \begin{tabular}{lrrrrrrr}
    \toprule
    Source & $\Delta$ ARS & Fwd & Rev & Keying gap & Extreme & Midpoint & SD ratio \\
    \midrule
    Humans ($n{=}263$) & 0.00 & 0.00 & 0.00 & 2.00 & 0.27 & 0.21 & 1.00 \\
    \midrule
    gpt-5.4-mini (LLM) & +0.21 & +0.35 & +0.07 & 1.51 & 0.02 & 0.33 & 0.28 \\
    GPT-6 Luna (LLM) & +0.53 & $-$0.66 & +1.71 & 1.12 & 0.00 & 0.45 & 0.39 \\
    Median of 47 generative LLMs & +0.21 & $-$1.20 & +1.40 & 1.20 & 0.03 & 0.38 & 0.54 \\
    \midrule
    Solar Decide Flash & +3.54 & +3.41 & +3.68 & 1.05 & 0.20 & 0.35 & 0.35 \\
    Microsoft-Decision-1 & +2.68 & $-$0.85 & +6.20 & 0.21 & 0.06 & 0.27 & 0.48 \\
    Kev 4B & +1.57 & +0.18 & +2.95 & 0.96 & 0.00 & 0.43 & 0.26 \\
    GPT-6 Luna Decisions & +1.48 & +0.78 & +2.18 & 1.45 & 0.02 & 0.27 & 0.32 \\
    Tev1 4B & +1.40 & +0.72 & +2.08 & 1.38 & 0.05 & 0.33 & 0.44 \\
    Liquid d1 & +1.11 & +0.86 & +1.35 & 1.74 & 0.06 & 0.22 & 0.41 \\
    Decider v1.1 27B & +1.09 & +0.20 & +1.98 & 1.31 & 0.00 & 0.34 & 0.24 \\
    Decider v1 27B & +1.08 & +0.18 & +1.99 & 1.31 & 0.00 & 0.34 & 0.24 \\
    Jev 1.13 & +1.01 & +1.10 & +0.92 & 1.72 & 0.08 & 0.29 & 0.46 \\
    Mercury Decide & +0.78 & +0.88 & +0.68 & 1.63 & 0.13 & 0.33 & 0.59 \\
    Drex v1.5 & +0.78 & $-$0.35 & +1.91 & 1.08 & 0.01 & 0.30 & 0.45 \\
    Clef & +0.32 & +0.23 & +0.40 & 1.72 & 0.00 & 0.22 & 0.28 \\
    Clef Flash & $-$0.23 & $-$0.99 & +0.53 & 1.28 & 0.01 & 0.42 & 0.46 \\
    Clef Omni & $-$0.27 & $-$0.96 & +0.41 & 1.44 & 0.02 & 0.31 & 0.44 \\
    \bottomrule
  \end{tabular}}
\end{table}

All language models compress the scale (Table~\ref{tab:style}). The share of extreme answers is \vHumanExtreme{} for humans, \vGptExtreme{} for gpt-5.4-mini and at most \vDecExtremeMax{} for the decision models. Every model uses the midpoint at least as often as humans, and most use it clearly more. On the balanced acquiescence index the median generative LLM shifts by \vGenMedAcq{} human SD and decision models by \vDecAcqMin{} to \vDecAcqMax{}. All decision models except Clef Flash and Clef Omni shift more than gpt-5.4-mini (\vGptAcqFair). For Kev 4B, both Decider versions, Drex v1.5 and Microsoft-Decision-1 almost all of this shift comes from the reverse-keyed items, and for Tev1 4B and GPT-6 Luna Decisions most of it does (columns Fwd and Rev). Their answers to reverse-keyed statements agree more than human answers do, while their forward-item means are closer to the human means. This pattern fits misresponse to reversed and negated items better than a general tendency to agree~\citep{weijters2012misresponse}. Jev~1.13, Liquid d1, Mercury Decide and Solar Decide Flash agree more than humans on both kinds of item, Solar Decide Flash by more than three human SD. Microsoft-Decision-1 is an extreme case of misresponse. Its reverse-item mean lies 6.2 human SD above the human value, and its keying gap is 0.21 against 2.00 in humans. On DUN\_02 (``I usually resist new ideas'') its mean expected score is 3.86, while the human mean is 1.86. It thus answers most negated statements as if they were worded positively. It is also the only model whose answers to reverse-keyed items are closer to the human answers before alignment than after (Appendix Table~\ref{tab:keying}). Every model has a smaller keying gap than humans, which means that models separate forward and reverse-keyed wording less sharply than people do.

\begin{table}[t]
  \centering\small
  \caption{Downstream use (C3, $n{=}100$, rounded readout). $R^2$: an OLS regression of performance on attitudes and engagement is fit on the synthetic sample and tested on a fixed 30\% split of the humans; the same regression fit on the other 70\% of humans reaches $R^2{=}0.28$. The raw column repeats the test without aligning reverse-keyed items, which is how the original audit scored it. Null paths: number of the 7 mediation triples with no human indirect effect for which the synthetic sample shows a significant indirect effect above 0.10. Of the 10 triples drawn, 3 are left out because a total score in them contains another of its variables.}
  \label{tab:downstream}
  \begin{tabular}{lrrrr}
    \toprule
    Source & $R^2$ (aligned) & $\Delta R^2$ & $R^2$ (raw) & Null paths \\
    \midrule
    Gaussian copula & 0.29 & +0.01 &  & 0/7 \\
    gpt-5.4-mini (LLM) & $-$1.79 & $-$2.07 & $-$0.61 & 6/7 \\
    MiniMax M3 (best LLM) & $-$0.24 & $-$0.51 & & 1/7 \\
    GPT-6 Luna (LLM) & $-$0.26 & $-$0.54 & & 2/7 \\
    Median of 47 generative LLMs & 0.10 & $-$0.18 & & 2/7 \\
    \midrule
    Jev 1.13 & 0.30 & +0.02 & 0.03 & 3/7 \\
    Decider v1 27B & 0.19 & $-$0.09 & $-$0.05 & 0/7 \\
    Decider v1.1 27B & 0.15 & $-$0.12 & $-$0.06 & 0/7 \\
    Liquid d1 & $-$0.61 & $-$0.89 & $-$1.31 & 5/7 \\
    Kev 4B & $-$0.14 & $-$0.42 & $-$0.19 & 2/7 \\
    Tev1 4B & $-$0.13 & $-$0.40 & $-$0.39 & 4/7 \\
    Mercury Decide & $-$0.06 & $-$0.33 & 0.07 & 2/7 \\
    Solar Decide Flash & $-$0.81 & $-$1.09 & $-$0.45 & 4/7 \\
    Clef & $-$0.42 & $-$0.70 & $-$0.18 & 1/7 \\
    Clef Flash & $-$0.47 & $-$0.75 & $-$0.51 & 6/7 \\
    Clef Omni & $-$0.95 & $-$1.22 & $-$1.16 & 6/7 \\
    GPT-6 Luna Decisions & 0.18 & $-$0.10 & 0.08 & 1/7 \\
    Microsoft-Decision-1 & $-$1.44 & $-$1.72 & $-$0.67 & 3/7 \\
    Drex v1.5 & $-$0.01 & $-$0.29 & $-$0.18 & 4/7 \\
    \bottomrule
  \end{tabular}
\end{table}

Table~\ref{tab:downstream} reports two tests of practical use. A regression trained on the copula sample predicts held-out humans as well as one trained on humans ($R^2$ \vCopulaRtwo{} against \vHumanRtwo). Among decision models only Jev matches this level (\vJevRtwo). Decider v1 (0.19), GPT-6 Luna Decisions (\vLunaRtwo) and Decider v1.1 (0.15) are next. Generative LLMs vary widely. Only \vGenRtwoPos{} of \vNGen{} beat predicting the mean, the median is \vGenMedRtwo{}, and \vGenRtwoNear{} reach 0.25 or more, with a maximum of \vGenRtwoMax{}. The aligned gpt-5.4-mini sample gives $R^2{=}\vGptRtwo$. In this sample attitudes and performance correlate at \vGptAttPerf{}, against \vHumanAttPerf{} in humans. Attitudes also predict performance more strongly than engagement does (\vGptEngPerf{} for engagement), which is the reverse of the human order (\vHumanEngPerf{}). A regression fit on this sample gives attitudes far too much weight for real people. This test uses a single human split and 100 training rows, so differences of less than about 0.1 should not be over-read.

Fabricated structure is common in the null-mediation test. The copula produces no false paths. Of the \vNullN{} triples, generative LLMs fabricate between \vGenNullMin{} and \vGenNullMax{} (median \vGenMedNull), gpt-5.4-mini \vGptNull{} and decision models between \vDecNullMin{} and \vDecNullMax{}. No false path appears for \vGenNullZero{} generative models and for both Decider versions. Clef and GPT-6 Luna Decisions show one each. The intervals of these four models are as narrow as those of the other models, so their low counts are not a matter of low power. At $n{=}100$ a single triple can change the count, and the ranking between models should be read with this in mind. Appendix~\ref{app:agreement} reports the agreement between models and the sensitivity of the ranking to the PSS weights.

\subsection{The four hypotheses}
\label{sec:res-hypotheses}
The probability vectors support the first hypothesis. Averaged vectors come much closer to the human marginals than any rounded answers. Sampled answers reach distribution components of 0.83--0.96, above most generative models (\vGenDistMin--\vGenDistMax), and this is the fair comparison because both sides are then samples of answers. With the rounded readout, only \vDecDistAboveGenMed{} of \vNDecision{} decision models exceed the generative median (\vGenMedDist).

The second hypothesis also holds. No decision model under a deterministic readout exceeds a correlation component of \vDecCorrMax{}, and \vGenAboveDecCorr{} of the \vNGen{} generative cells are above this value. This follows partly from the format, as the profile regression shows (Section~\ref{sec:res-readouts}). Decision models also reproduce the factor structure less well, because their person signal has nearly one direction (Section~\ref{sec:res-structure}).

Only part of the third hypothesis holds. A small share of each vector depends on the respondent, but this share is larger than the profile supports, and the vectors rank respondents about as well as a regression on the profile. Without a profile, the sampled PSS of decision models falls by \vGainMin{} to \vGainMax. The problem is calibration rather than a weak use of the profile.

Finally, the comparison of release waves does not support the fourth hypothesis (Section~\ref{sec:res-newer}).

\section{Discussion}
\label{sec:discussion}
\paragraph{Marginals and cost.} Decision models remove the practical problems of free-text generation and cost a fraction of a cent per respondent. They also return a full answer distribution. Averaged over a sample, this distribution estimates item marginals better than the rounded answers of any language model in our set. On one fixed human split, a regression trained on Jev's synthetic sample predicts held-out humans about as well as one trained on humans. Only \vGenRtwoNear{} of \vNGen{} generative models come close to this.

\paragraph{Correlation structure and calibration.} Decision models reproduce the pattern of item correlations and the factor structure less well than most generative LLMs. Their person signal moves nearly all items in one direction, so the subscales are less distinct than in human data. They do read the profile, and their vectors rank people about as well as a regression on the profile. However, their average vector differs from the answer frequencies of real respondents, and they move each respondent further from the population than the profile supports. For these reasons their forecasts of single people are worse than the answer frequencies of other people until they are shrunk toward these frequencies. The same loss of correlation structure appears within one model name, as GPT-6 Luna loses much of it when served as a decision model. In most decision models acquiescence is also higher than in generative models, mainly on reverse-keyed items, and all of them compress the scale. All decision models except the two Decider versions still show at least one indirect effect that humans do not show. Therefore, a cleaner output format does not fix the problems that the original audit found in generative samples. These problems come from the way the model represents a respondent, and the output format changes only which of them are visible.

\paragraph{How much a profile can tell.} A regression that sees the same profile fields and the answers of 163 other people improves the forecast of a person's answers by only \vFcRegGain\%. Most of the variation in how people answer these items is not explained by the profile. \citet{hu2024quantifying} found the same for annotation tasks, where persona variables explain less than 10\% of the variance in most datasets. A model that individualises strongly from these fields will be wrong more often, as the narrow repeat distributions of the generative models and the over-spread decision vectors show. One reading of the correlation results follows from this. The dependence between items in human data comes from differences between people that the profile does not record. A synthetic respondent can reproduce it only by inventing such differences. Generative models do so and gain on the correlation component, but they overdo it. Their items correlate more strongly than in human data, and one general factor carries more of the variance. Decision models stay closer to the population and lose on the correlation component. For profile-based synthetic respondents, the realistic goal is therefore fidelity at the level of the sample.

\paragraph{Newer releases.} We expected newer models to do at least as well as their predecessors. Instead, the newest release scores lower in \vSuccDown{} of \vSuccN{} product lines, and in \vSuccBelow{} of them the drop is larger than the noise. The new models also answer with less spread, although the comparisons of release waves are exploratory. The one decision-model successor, Decider v1.1 27B, gives almost the same answers as its predecessor. We did not test why the generative releases lose fidelity. One possible reason is post-training that favours moderate answers in assistant use, which would push a model away from the spread of real survey responses. Reinforcement learning from human feedback is known to reduce the diversity of model outputs~\citep{kirk2024understanding}, which fits this explanation. A model upgrade therefore cannot be assumed to preserve the fidelity of a synthetic sample, and a study that switches to the newest version may lose fidelity without noticing.

\paragraph{Recommendations for evaluating synthetic respondents.} The results give nine recommendations. First, check the orientation of reverse-keyed items on both sides of every comparison. One item with a known answer, such as DUN\_02, is enough to catch a mismatch. Second, check that the analysis code selects the right rows for every kind of source, including baselines, because a missing key can drop a component without an error message. Third, report the components with standard errors next to any weighted total. At $n{=}100$ the mediation component alone can move PSS by 0.1. Fourth, treat the readout as part of the method and fix it before scoring, because it decides which property of the sample is kept. Fifth, validate each model version separately, because a newer release can score lower. Sixth, express counterfactual effects in units of the human SD. The paired $d$ also depends on how noisy a model is, so it can rank a deterministic model above a noisier one with a larger shift. Seventh, when a model returns probabilities, score them against each person's own answers with a proper scoring rule. Compare them with the answer frequencies of other respondents, with the model's own average forecast and with a recalibrated version. Eighth, compare synthetic samples with a second human sample of the same size. It shows how much of a similarity score is sampling noise. Ninth, report the level of the item correlations and the factor structure next to any similarity of correlation patterns, because a pattern can match while all correlations are too strong.

\paragraph{Advice for survey practice.} If an analyst needs rough item marginals before data collection, the vector of a decision model averaged over real profiles of the target population is a cheap first estimate. Without such profiles it is further from the human marginals for most models (Section~\ref{sec:res-controls}). Even with them, it forecasts answers worse than the answer frequencies of other respondents, so it replaces neither a pilot study nor a pretest. In the rounded answers of decision models the median item SD is \vDecSdRatioMin--\vDecSdRatioMax{} of the human value. Their answers should therefore not be used to plan sample sizes or to check reliability. In terms of total survey error, the error of the model is a bias. It does not shrink when more synthetic respondents are added, and standard errors computed from a synthetic sample do not include it. Our human reference is a non-probability online sample, so all fidelity results refer to this sample. Where real data from a similar population exist, a copula fit to them is better. For analyses of relations between constructs, neither class of language model is a substitute for a human sample. No model in our set predicts how a particular person will answer better than the answer frequencies of other people.

\section{Limitations and future work}
\label{sec:limitations}
\paragraph{Dataset and conditions.} All results come from one instrument set, in one language and one country, and from self-reports. We use one Lithuanian sample, 100 respondents, the C3 profile and, for the decision models, a control without a profile. The intermediate conditions of the profile ladder in the original audit were not repeated. GPT-6 Luna Decisions returned HTTP 502 for every request with an empty state, so it has no C0 cell. Decider v1 27B has no role and gender swaps because its endpoint was unavailable when they were run. The observed human education contrast rests on 15 secondary-educated respondents and is too noisy to say whether any model exaggerates it.

\paragraph{Selection among many cells.} With \vNGen{} generative cells, the best one is partly selected by chance. We report paired intervals against it and against the fixed reference gpt-5.4-mini, but the ranking inside the top group is not stable. A second sample of respondents would be needed to confirm the order of the leading models.

\paragraph{Noise at $n{=}100$.} The mediation and demographic components are noisy at this sample size. A second sample of 100 real people scores below the best language models on PSS when both are compared with the headline sample (Section~\ref{sec:res-field}). At this size PSS can separate weak from strong synthetic samples, but it cannot rank the best models against real respondents. A run on all 263 profiles would narrow the intervals for the models that matter most. With \vNDraws{} bootstrap draws the interval ends carry some Monte Carlo error, which matters only for differences close to zero.

\paragraph{Reasoning setting and timing.} All generative cells use minimal reasoning. For \vNHigh{} models the original audit also ran high reasoning effort. With it, \vNHighUp{} of them score higher and the others lower, so the setting can move a model in either direction. Newer models may also be tuned for longer reasoning than the minimal setting allows. The new models were queried in October 2026, months after the older cells, and we did not re-query the older models. Vendor-side changes to them since then are not captured. The new open-weight models and Grok 4.7 were also reached through different providers from those in the original audit (Fireworks and the xAI API instead of the Nexos gateway).

\paragraph{Request design.} The profile is passed to decision models in English JSON while the items and anchors are Lithuanian. Some models may handle the mixed language worse than a fully Lithuanian request. The three Clef models and Drex v1.5 answer the last four items in a separate request that does not see the first 64. Tev1 4B gives a check of this split. Its unchanged profiles, sent again in three parts, repeat \vTevFormatSame\% of its rounded answers (total variation \vTevFormatTv), which is the size of its run-to-run noise.

\paragraph{Moving targets.} The decisions endpoint is an alpha API. We record the resolved model versions, but the endpoint exposes no seed. Our retest repeats 20 respondents once on the same day, so it does not show drift over weeks. Solar Decide is missing because of its latency and server errors, but its low-latency variant Solar Decide Flash is included.

\paragraph{Structure tests.} Tucker's $\varphi$ uses an exploratory solution with greedy factor alignment. We did not test measurement invariance with multi-group confirmatory models. Several decision samples contain items without any variance (13 for each Decider version), and at $n{=}100$ such models are unstable.

\paragraph{Memorisation.} The human data have been public since 2020. We did not test the decision models for memorisation, and their close match to the item marginals could partly reflect it. However, for \vJsWorse{} of the \vNCzero{} models with a C0 grid the vector without a profile is further from the human marginals than the vector averaged over real profiles. This argues against a simple recall of the marginals.

\paragraph{Forecast references.} The population distribution and the profile regression are fit on 163 people from the same survey, which no language model sees. Generative forecasts come from 20 repeats of 26 to 30 respondents at a requested temperature of 0.7 (the provider default where the API fixes it), and a higher temperature would widen them. These results thus compare the models with what can be done with data from the same population. They do not show how far each model is from the best possible forecast.

\paragraph{Scope of the scoring update.} This paper recomputes the headline numbers, the ranking, response style, predictive validity and the null-mediation test at C3. The other analyses of the original audit, such as the profile ladder and the generative swap grids, are recomputed in its revised version.

\paragraph{Future work.} The results point to several directions. First, the profile fields improve the forecast of a person's answers by only \vFcRegGain\%. Future studies should give the model richer information about each person, such as answers to a few anchor items, answers from an earlier wave or an interview. They can then test whether the forecasts beat the answer frequencies of other respondents. Second, mixing each vector with the population distribution removed most of the forecast loss of every decision model. Recalibration of this kind, fit on a small human sample, should become a standard step before the vectors are used. It remains open how many real respondents such a calibration needs, and at which pilot size a copula fit on the same pilot is better. Third, a decision request scores each item without a shared respondent state, and the person signal of the vectors has nearly one direction. Requests for the joint distribution of item pairs, or sequential requests that condition each item on earlier answers, would show whether the weak factor structure comes from the interface or from the model. Fourth, the two classes fail on different components. A hybrid that combines sampled decision marginals with the correlation pattern of generative samples could be tested without any human answers. Fifth, all results come from one Lithuanian sample and four instruments. Replications in other languages, with other instruments and with probability samples are needed, in particular for the handling of reverse-keyed items. Sixth, in \vSuccBelow{} of \vSuccN{} product lines the newest release scored lower beyond noise. A fixed benchmark of this kind can serve as a regression test that is run again whenever a provider changes a model version. Finally, at $n{=}100$ PSS does not separate the best language models from a second sample of real people. Runs on all 263 profiles, or on a larger survey, are needed before models can be ranked against real respondents.

\section{Conclusion}
\label{sec:conclusion}
This paper aimed to test whether decision models, or newer generative LLMs, are better synthetic survey respondents than earlier generative LLMs. On an open psychometric benchmark, neither class does better. Decision models are a cheap and clean interface, and their probability vectors match the answers of the population closely. On the other hand, they recover less item dependence than most generative LLMs and react to role less strongly than generative models and people (median \vRoleShiftMed{} against \vGenRoleShiftMed{} and \vHumanRoleD{} human SD). They also spread respondents too far apart. For this reason they forecast single people worse than the answer frequencies of other people, although they rank people about as well as a regression on the profile. The readout that turns a vector into one answer decides which property of the sample is kept. Newer generative models do not close the gap, and in most product lines the newest release scores below the one before it.

A statistical model of real responses fit on other people matches the best of \vNGen{} language models, and the same model fit in sample beats all of them. The revised scoring changes the order of the LLMs and makes the main conclusion of the original audit stronger. On the full reference the real answers of the same people also score higher than any language model. A second human sample of the same size shows, however, that at $n{=}100$ PSS cannot separate the best language models from real respondents. Finally, the profile fields explain little of how a person answers. Work on profile-based synthetic respondents should therefore aim at fidelity at the level of the sample. Richer information about each person would be needed before predicting a single person becomes a realistic goal.

\paragraph{Ethics statement.} The human data were collected in 2020 for a master's thesis by the second author~\citep{sarkauskaite2020}, under the informed-consent protocol of the awarding university, which covered academic re-use of de-identified data. No direct identifiers were collected, and no new human data were collected for this study. Each model request contained only the eleven coarse profile fields and no identifiers or free text from respondents. Mercury Decide was called through a free endpoint whose provider may retain prompts.

\paragraph{Data and code availability.} The de-identified human data are those of the original audit (CC BY-NC 4.0). The code, the raw model outputs with the full probability vectors, and the analysis tables are available from the corresponding author on reasonable request. Appendix~\ref{app:repro} describes the pipeline.

\bibliographystyle{plainnat}
\bibliography{references}

\clearpage
\appendix

\section{Example request and response}
\label{app:request}
The state for one respondent (field values as stored in the processed profile file):
\begin{small}
\begin{verbatim}
{"country": "Lithuania", "gender": "woman", "age": "22",
 "education": "university (Bachelor/Master/Doctorate)",
 "position": "non-managerial", "tenure_total": "less than 1 year",
 "tenure_current_org": "less than 6 months", "org_sector": "services",
 "org_size": "small (up to 50 employees)",
 "change_intensity": "9", "change_relevance": "6"}
\end{verbatim}
\end{small}
One of the 68 score questions (item IWPQ\_01). The instruction is ``Atsakyk taip, kaip \v{s}is darbuotojas atsakyt\k{u} \k{i} teigin\k{i}.'' followed by the item text ``A\v{s} gebu suplanuoti savo darb\k{a} taip, kad j\k{i} baig\v{c}iau laiku'' (``I can plan my work so that I finish it on time''). The criteria are ``1 Niekada'', ``2 Retai'', ``3 Kartais'', ``4 Da\v{z}nai'' and ``5 Visuomet'' (never, rarely, sometimes, often, always). Jev~1.13 returned the probabilities $(0.01, 0.15, 0.53, 0.30, 0.01)$ for levels 1--5. The expected score is 3.15, so both the rounded readout and the most probable category give 3.

\section{Scoring update relative to the published version of the original audit}
\label{app:scoring}
\begin{figure}[t]
  \centering
  \includegraphics[width=0.7\linewidth]{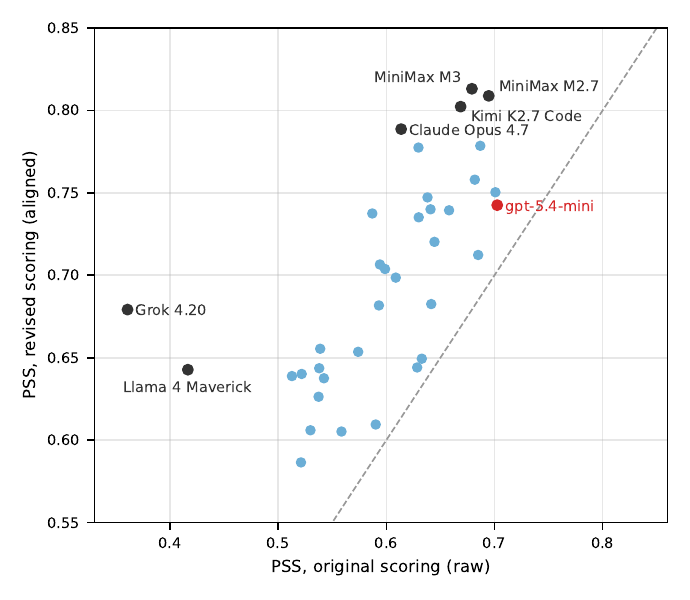}
  \caption{PSS of the \vNGenFirst{} C3 cells of the original audit under the published scoring (raw) and the revised scoring (aligned). The dashed line marks equal scores. gpt-5.4-mini, the published leader, is shown in red.}
  \label{fig:rankchange}
\end{figure}

While re-running the benchmark we revised two steps of the original scoring and one style index. The revised arXiv version of the original audit~\citep{lukauskas2026plausible} uses the same code. Table~\ref{tab:corrections} lists the numbers that change relative to the published version, and Figure~\ref{fig:rankchange} shows the change in the ranking. Here every sample is scored on the same 100 respondents. For this reason some values differ from the revised version of the original audit in the third decimal. The in-sample copula differs more, because that paper scores it on all 263 draws (PSS 0.936).

\begin{table}[!htbp]
  \centering\small
  \caption{Numbers from the first version (v1) of the original audit~\citep{lukauskas2026plausible} that change under the revised scoring, computed on the same 100 respondents. The revised version (v2) of that paper uses the same code. Causes: (A) model answers to reverse-keyed items were not reversed before scoring; (B) the distribution component was missing from the baseline totals; (C) the published index measured keying consistency rather than agreement; (D) the published test pooled all LLM cells; (E) null triples in which a total score contains another variable of the triple are left out. $^{\dagger}$Qwen 3.7 Plus has fewer than 90 complete respondents at C3 and is not ranked. $^{\ast}$Pooled over 37 models and all conditions ($n{=}24{,}267$).}
  \label{tab:corrections}
  \resizebox{\linewidth}{!}{%
  \begin{tabular}{lllc}
    \toprule
    Quantity & Original audit, v1 & Revised & Cause \\
    \midrule
    PSS, gpt-5.4-mini (C3) & 0.714 & 0.743 & A \\
    Rank of gpt-5.4-mini among the models of the original audit & 1 of 37 & 10 of 36$^{\dagger}$ & A \\
    Best LLM of the original audit & gpt-5.4-mini, 0.714 & MiniMax M3, 0.813 & A \\
    Correlation component, gpt-5.4-mini & 0.522 & 0.602 & A \\
    PSS, Gaussian copula & 0.688 & 0.927 (in-sample), 0.824 (out-of-sample) & B \\
    In-sample copula minus best LLM & $-$0.026 & +0.114 & A, B \\
    Acquiescence shift, gpt-5.4-mini & +1.29 & +0.21 SD & A, C \\
    Synthetic-trained $R^2$ on humans, gpt-5.4-mini & $-$0.62 & $-$1.79 & A \\
    Null paths with a fabricated indirect effect & 3/10$^{\ast}$ & 6/7 (gpt-5.4-mini) & A, D, E \\
    \bottomrule
  \end{tabular}}
\end{table}

\paragraph{Reverse-keyed items.} Human answers to the 13 reverse-keyed items are stored reversed, so that a high value always means a positive attitude. Model answers to these items were stored as given to the original wording and were not reversed before scoring. We now reverse them with $x' = \min + \max - x$ and call the result the aligned score. The published scores are raw. A consistent respondent with a positive attitude therefore scored high on a reverse-keyed item in the human file and low in the model file. Item DUN\_02 (``I usually resist new ideas'') shows the size of the effect. The human mean is 4.14 after reversal, while gpt-5.4-mini answers 1.97 on average, which is close to the human mean before reversal (1.86). The change raises the PSS of every one of the \vNGenFirst{} C3 cells of the original audit, by a median of \vMedGain{} and at least \vMinGain{}. For gpt-5.4-mini the aligned PSS is \vGptTotal{}. The change affects all five components. It removes divergence on 13 items, restores the sign of their correlations with the other 55 items, changes the Dunham $\alpha$, and enters the attitude composites used for mediation and demographic contrasts. Table~\ref{tab:keying} shows that the raw comparison roughly doubles the divergence on reverse-keyed items. Our raw recomputation of gpt-5.4-mini gives \vGptRawTotal{}, against the published 0.714. The gap comes from pooling. For most models the published version averaged the 20 test-retest repeats into the C3 cell, while we use the single run of the main grid for every model. We also checked whether the gain of each model follows its keying gap. The rank correlation is \vGainKeyRho{} ($p{=}\vGainKeyP$), so the gain depends on more than this one property.

\paragraph{Baseline totals.} The distribution component was missing from the totals of the statistical baselines and counted as zero. With it included, the copula of the original audit reaches \vCopulaTotal{} on the same 100 IDs. This copula is fit on the 263 people it is scored against, so we also report an out-of-sample copula, fit on the 163 people outside the headline sample, which reaches \vCopulaOosTotal{}. The analysis script computes item distances for every cell and selects the rows of one cell by matching model, condition, presentation mode and reasoning effort. Baseline samples have no reasoning-effort value, and a missing value never matches itself. The selection was empty, and the total treated the missing component as zero. The published copula total of 0.688 is the weighted sum of the other four components alone.

\paragraph{Acquiescence.} The index of the original audit, the forward-item mean minus the reverse-item mean, grows with answers that follow the content of the items. We keep it under the name keying gap and add the balanced acquiescence index (ARS)~\citep{baumgartner2001response}. The keying gap gives zero to a respondent who agrees with every statement and a large positive value to a respondent who follows the content of the items. ARS is the average of the forward-item mean and the reverse-item mean on a 0--1 scale, minus 0.5, computed on raw answers. With ARS computed in this way for humans and models, gpt-5.4-mini shifts by \vGptAcqFair{} human SD. Humans have a larger keying gap (\vHumanKeying) than gpt-5.4-mini (\vGptKeying). Applying the keying gap to reversed human answers and unreversed model answers reproduces the published shift for gpt-5.4-mini (\vGptAcqPublishedStyle{} against the published +1.29).

\paragraph{Ranking.} The first change raises some models more than others, so the order of the LLMs changes (Figure~\ref{fig:rankchange}). The Spearman correlation between the published and the revised order is \vRankRho{}. gpt-5.4-mini moves from first to \vGptRankFirst{}th place, and \vBestGenName{} (\vBestGenTotal), \vSecondGenName{} (\vSecondGenTotal) and \vThirdGenName{} (\vThirdGenTotal) now lead. The correlation component gains most and rises for every cell (median \vGainCorrMed). The reliability component also rises for every cell (median \vGainRelMed), because the unreversed items had moved the Dunham $\alpha$ of some models far from the human value. The mediation component falls for \vGainMediDown{} of \vNGenFirst{} cells (median \vGainMediMed). The order changes mainly through the reliability and mediation components, whose changes differ most between models (SD \vGainRelSd{} and \vGainMediSd, against \vGainCorrSd{} for correlation).

\paragraph{What holds and what changes.} The main conclusion of the original audit holds and becomes stronger. LLM samples are not a drop-in replacement for human data, and the lead of the statistical baseline grows. In its published version the best LLM was slightly above the copula. With the revised scoring the in-sample copula is above every language model, by \vBCopulaTotal{} [\vBCopulaTotalLo, \vBCopulaTotalHi] over the best one (Table~\ref{tab:paired}), and the out-of-sample copula is level with the best (\vBCopulaOosTotal{} [\vBCopulaOosTotalLo, \vBCopulaOosTotalHi]). Range restriction holds as well, because it does not depend on item direction. The share of extreme answers is \vHumanExtreme{} for humans and \vGptExtreme{} for gpt-5.4-mini. LLM samples still produce mediation paths that humans do not show (Section~\ref{sec:res-downstream}). The revision changes three findings. First, the best LLM is a different model. Second, the acquiescence finding becomes a pattern of fewer extreme answers, more midpoint answers and a smaller keying gap. Third, the advantage of LLMs on mediation and demographic effects shrinks, as the next paragraph shows.

\paragraph{Mediation and demographic effects.} The original audit argued that these components are where language models add value over statistical baselines. Against the in-sample copula this does not hold. Its mediation component is \vCopulaMed{} against \vGptMed{} for gpt-5.4-mini, and its demographic component is \vCopulaDemo{} against \vGptDemo{} (Table~\ref{tab:leaderboard}). Against the out-of-sample copula the mediation component is \vCopulaOosMed{}, and the paired difference to gpt-5.4-mini is \vDCopulaOosMed{} [\vDCopulaOosMedLo, \vDCopulaOosMedHi]. Its demographic component is \vCopulaOosDemo{}, with a difference of \vDCopulaOosDemo{} [\vDCopulaOosDemoLo, \vDCopulaOosDemoHi], which is within noise. Neither copula sees the profile, so its demographic component reflects chance agreement only, and the gap between the two copulas shows how noisy this component is at $n{=}100$. PSS scores only one indirect effect, and gpt-5.4-mini reproduces it better than the real answers of the same 100 people do (mediation component \vGptMed{} against \vOracleMed). The same sample produces \vGptNull{} of the \vNullN{} cross-construct mediation paths that are absent in the human data.

\paragraph{Analyses not affected.} Results that compare models only with each other, such as inter-LLM similarity and test-retest reliability, do not depend on the direction of reverse-keyed items. The revised version of the original audit recomputes the analyses that compare models with humans through these items. It also signs the paired differences of its counterfactual swaps by the direction of the swap and reports the swap effects in units of the human SD. Without the sign, swaps in opposite directions cancel, which made the role effect look small.

\section{Models, costs and item keying}
\label{app:models}
Table~\ref{tab:models} lists the decision models with their pinned versions, context limits, prices and the cost of the C3 grid. Table~\ref{tab:keying} compares the item divergence of reverse-keyed and forward-keyed items under the raw and the aligned scoring.

\begin{table}[!htbp]
  \centering\small
  \caption{Decision models in the study. All were called through the OpenRouter decisions endpoint with one score question per item. \emph{Calls} is the number of requests per respondent: the three Clef models and Drex accept at most 64 questions, so they receive the profile twice (64 items, then 4). \emph{Price} is the listed input price in USD per million tokens; output tokens are free for this endpoint. \emph{Cost} is the observed mean cost per synthetic respondent in thousandths of a dollar. The C3 grid of the scored models cost \$1.27, and the incomplete Solar Decide run another \$0.08.}
  \label{tab:models}
  \resizebox{\linewidth}{!}{%
  \begin{tabular}{llrrrrr}
    \toprule
    Model & Resolved version & Calls & $n$ & Input tokens & Price & Cost ($10^{-3}$\,\$) \\
    \midrule
    Jev 1.13 & \texttt{typesafe/jev-1.13-20260917} & 1 & 100 & 7,800 & 0.042 & 0.33 \\
    Decider v1 27B & \texttt{perplexity/pplx-decider-v1-27b-20261001} & 1 & 100 & 17,817 & 0.040 & 0.71 \\
    Decider v1.1 27B & \texttt{perplexity/pplx-decider-v1.1-27b-20261006} & 1 & 100 & 17,817 & 0.020 & 0.36 \\
    Liquid d1 & \texttt{liquid/d1-20260930} & 1 & 100 & 15,953 & 0.040 & 0.64 \\
    Kev 4B & \texttt{jaredpalmer/kev-4b-20260924} & 1 & 100 & 5,607 & 0.042 & 0.24 \\
    Tev1 4B & \texttt{togethercomputer/tev1-4b-experimental-20260923} & 1 & 100 & 18,264 & 0.042 & 0.77 \\
    Mercury Decide & \texttt{inception/mercury-decide-20260930} & 1 & 100 & 13,230 & 0.000 & 0.00 \\
    Solar Decide\,$^{\dagger}$ & \texttt{upstage/solar-decide-20260928} & 1 & 46 & 35,091 & 0.050 & 1.75 \\
    Solar Decide Flash & \texttt{upstage/solar-decide-flash-20261008} & 1 & 100 & 35,044 & 0.050 & 1.75 \\
    Clef & \texttt{cloudflare/clef} & 2 & 100 & 12,635 & 0.240 & 3.03 \\
    Clef Flash & \texttt{cloudflare/clef-flash} & 2 & 100 & 12,635 & 0.090 & 1.14 \\
    Clef Omni & \texttt{cloudflare/clef-omni-20261009} & 2 & 100 & 13,505 & 0.150 & 2.03 \\
    GPT-6 Luna Decisions & \texttt{openai/gpt-6-luna-decisions-20261006} & 1 & 100 & 12,025 & 0.100 & 1.20 \\
    Microsoft-Decision-1 & \texttt{microsoft/microsoft-decision-1-20261009} & 1 & 100 & 6,140 & 0.042 & 0.26 \\
    Drex v1.5 & \texttt{nace-ai/drex-v1.5-20261009} & 2 & 100 & 5,693 & 0.040 & 0.23 \\
    \bottomrule
  \end{tabular}}
  \par\smallskip\raggedright\footnotesize $^{\dagger}$\,Incomplete at the time of writing; not scored.
\end{table}

\begin{table}[!htbp]
  \centering\small
  \caption{Mean item Jensen-Shannon divergence to the human marginal on the two instruments with reverse-keyed items (rounded readout). The raw column compares unflipped model answers with reverse-coded human answers, which is what the original audit did for every reverse-keyed item.}
  \label{tab:keying}
  \begin{tabular}{lrrr}
    \toprule
    Source & Forward items & Reverse items (aligned) & Reverse items (raw) \\
    \midrule
    gpt-5.4-mini (LLM) & 0.243 & 0.356 & 0.728 \\
    Jev 1.13 & 0.132 & 0.173 & 0.524 \\
    Decider v1 27B & 0.263 & 0.411 & 0.602 \\
    Decider v1.1 27B & 0.265 & 0.413 & 0.604 \\
    Liquid d1 & 0.168 & 0.288 & 0.561 \\
    Kev 4B & 0.248 & 0.470 & 0.555 \\
    Tev1 4B & 0.194 & 0.367 & 0.548 \\
    Mercury Decide & 0.091 & 0.139 & 0.467 \\
    Solar Decide Flash & 0.351 & 0.449 & 0.504 \\
    Clef & 0.223 & 0.352 & 0.681 \\
    Clef Flash & 0.184 & 0.282 & 0.606 \\
    Clef Omni & 0.206 & 0.275 & 0.615 \\
    GPT-6 Luna Decisions & 0.225 & 0.363 & 0.553 \\
    Microsoft-Decision-1 & 0.145 & 0.551 & 0.291 \\
    Drex v1.5 & 0.170 & 0.306 & 0.508 \\
    \bottomrule
  \end{tabular}
\end{table}

\section{All generative cells}
\label{app:generative}
Table~\ref{tab:generative} lists the PSS components of all \vNGen{} generative cells under the revised scoring, with the raw PSS for the cells of the original audit.

\begin{table}[p]
  \centering\scriptsize
  \caption{All 47 generative cells at C3 (single call, minimal reasoning, same 100 respondents), ordered by aligned PSS. $^{\ast}$Released after the original audit and run for this study (11 models). Raw PSS is the original scoring path and Raw rank its order among the models of the original audit. The aligned columns are the components, the total and its bootstrap SE. Null is the number of the 7 null mediation triples with a fabricated path (Table~\ref{tab:downstream}). Qwen 3.7 Plus (75 complete respondents) and the six cells run with high reasoning effort are not shown.}
  \label{tab:generative}
  \setlength{\tabcolsep}{3pt}
  \resizebox{\linewidth}{!}{%
  \begin{tabular}{rllrrrrrrrrrrr}
    \toprule
    & & & & \multicolumn{2}{c}{Raw} & \multicolumn{7}{c}{Aligned} & \\
    \cmidrule(lr){5-6}\cmidrule(lr){7-13}
    \# & Model & Vendor & $n$ & PSS & Rank & Dist. & Corr. & Rel. & Med. & Demo. & PSS & SE & Null \\
    \midrule
    1 & MiniMax M3 & Open weights & 100 & 0.679 & 7 & 0.71 & 0.76 & 0.97 & 1.00 & 0.51 & 0.813 & 0.023 & 1 \\
    2 & MiniMax M2.7 & Open weights & 100 & 0.695 & 3 & 0.83 & 0.81 & 0.98 & 0.76 & 0.52 & 0.809 & 0.039 & 2 \\
    3 & Kimi K2.7 Code & Open weights & 100 & 0.669 & 8 & 0.80 & 0.81 & 0.95 & 0.80 & 0.52 & 0.802 & 0.030 & 1 \\
    4 & Claude Opus 4.7 & Anthropic & 100 & 0.614 & 18 & 0.72 & 0.77 & 0.96 & 0.96 & 0.34 & 0.789 & 0.033 & 2 \\
    5 & Gemini 3.1 Flash-Lite & Google & 100 & 0.687 & 4 & 0.86 & 0.72 & 0.95 & 0.75 & 0.42 & 0.779 & 0.047 & 5 \\
    6 & Qwen3.5 397B & Open weights & 100 & 0.630 & 16 & 0.76 & 0.65 & 0.97 & 0.82 & 0.65 & 0.777 & 0.033 & 1 \\
    7 & Gemini 2.5 Pro & Google & 100 & 0.682 & 6 & 0.87 & 0.76 & 0.94 & 0.57 & 0.48 & 0.758 & 0.042 & 0 \\
    8 & DeepSeek V4 Pro & Open weights & 100 & 0.701 & 2 & 0.80 & 0.80 & 0.95 & 0.49 & 0.61 & 0.750 & 0.051 & 1 \\
    9 & Kimi K2.6 & Open weights & 100 & 0.638 & 13 & 0.76 & 0.70 & 0.95 & 0.71 & 0.48 & 0.747 & 0.034 & 1 \\
    10 & gpt-5.4-mini & OpenAI & 100 & 0.703 & 1 & 0.66 & 0.60 & 0.96 & 0.93 & 0.49 & 0.743 & 0.020 & 6 \\
    11 & Claude Opus 4.8 & Anthropic & 100 & 0.641 & 12 & 0.70 & 0.74 & 0.96 & 0.79 & 0.29 & 0.740 & 0.038 & 2 \\
    12 & Claude Sonnet 5 & Anthropic & 100 & 0.658 & 9 & 0.70 & 0.70 & 0.96 & 0.77 & 0.41 & 0.739 & 0.039 & 3 \\
    13 & gpt-oss-120b & Open weights & 98 & 0.587 & 24 & 0.73 & 0.80 & 0.94 & 0.60 & 0.46 & 0.737 & 0.033 & 1 \\
    14 & Gemini 3.5 Flash & Google & 100 & 0.630 & 15 & 0.82 & 0.77 & 0.93 & 0.54 & 0.41 & 0.735 & 0.045 & 4 \\
    15 & Claude Sonnet 5.5$^{\ast}$ & Anthropic & 100 & 0.639 &  & 0.69 & 0.70 & 0.98 & 0.73 & 0.44 & 0.735 & 0.034 & 1 \\
    16 & Claude Haiku 4.5 & Anthropic & 100 & 0.644 & 10 & 0.75 & 0.73 & 0.95 & 0.59 & 0.43 & 0.720 & 0.034 & 2 \\
    17 & Qwen 3.8 Max$^{\ast}$ & Open weights & 100 & 0.641 &  & 0.71 & 0.66 & 0.99 & 0.66 & 0.48 & 0.720 & 0.031 & 0 \\
    18 & GLM 5.1 & Open weights & 100 & 0.685 & 5 & 0.82 & 0.78 & 0.95 & 0.39 & 0.46 & 0.712 & 0.053 & 1 \\
    19 & Gemini 3.8 Flash$^{\ast}$ & Google & 100 & 0.613 &  & 0.80 & 0.78 & 0.95 & 0.45 & 0.35 & 0.710 & 0.045 & 2 \\
    20 & Grok 4.3 & xAI & 100 & 0.594 & 21 & 0.72 & 0.76 & 0.96 & 0.48 & 0.50 & 0.707 & 0.048 & 3 \\
    21 & Nemotron 3 Super & Open weights & 99 & 0.599 & 20 & 0.83 & 0.76 & 0.96 & 0.31 & 0.53 & 0.704 & 0.035 & 2 \\
    22 & DeepSeek V4.1 Flash$^{\ast}$ & Open weights & 100 & 0.616 &  & 0.77 & 0.72 & 0.98 & 0.48 & 0.37 & 0.703 & 0.038 & 2 \\
    23 & Gemini 3 Flash & Google & 100 & 0.609 & 19 & 0.85 & 0.79 & 0.93 & 0.30 & 0.43 & 0.699 & 0.046 & 2 \\
    24 & GLM 5.3$^{\ast}$ & Open weights & 100 & 0.606 &  & 0.74 & 0.74 & 1.00 & 0.44 & 0.37 & 0.695 & 0.033 & 1 \\
    25 & Grok 4.20 (reasoning) & xAI & 100 & 0.641 & 11 & 0.75 & 0.74 & 0.96 & 0.41 & 0.36 & 0.683 & 0.049 & 1 \\
    26 & Claude Fable 5 & Anthropic & 100 & 0.593 & 22 & 0.75 & 0.72 & 0.96 & 0.46 & 0.33 & 0.682 & 0.034 & 2 \\
    27 & Grok 4.20 & xAI & 100 & 0.361 & 36 & 0.71 & 0.65 & 0.97 & 0.50 & 0.46 & 0.679 & 0.052 & 2 \\
    28 & DeepSeek 3.2 & Open weights & 100 & 0.539 & 28 & 0.87 & 0.75 & 0.94 & 0.12 & 0.39 & 0.655 & 0.040 & 0 \\
    29 & Gemini 3.1 Pro & Google & 100 & 0.574 & 25 & 0.83 & 0.78 & 0.95 & 0.10 & 0.39 & 0.654 & 0.035 & 3 \\
    30 & Gemma 4 31B & Open weights & 100 & 0.633 & 14 & 0.79 & 0.74 & 0.96 & 0.16 & 0.42 & 0.649 & 0.040 & 0 \\
    31 & GPT-5.4 nano & OpenAI & 100 & 0.628 & 17 & 0.66 & 0.55 & 0.92 & 0.56 & 0.46 & 0.644 & 0.039 & 0 \\
    32 & Grok 4.1 Fast (reasoning) & xAI & 100 & 0.538 & 29 & 0.90 & 0.73 & 0.93 & 0.00 & 0.50 & 0.644 & 0.027 & 3 \\
    33 & Llama 4 Maverick & Open weights & 100 & 0.416 & 35 & 0.70 & 0.57 & 0.97 & 0.43 & 0.45 & 0.643 & 0.038 & 1 \\
    34 & Kimi K3$^{\ast}$ & Open weights & 100 & 0.565 &  & 0.72 & 0.72 & 0.98 & 0.27 & 0.30 & 0.641 & 0.033 & 2 \\
    35 & Llama 4 Scout & Open weights & 100 & 0.522 & 32 & 0.73 & 0.63 & 0.98 & 0.48 & 0.08 & 0.640 & 0.030 & 5 \\
    36 & Grok 4.1 Fast & xAI & 100 & 0.513 & 34 & 0.90 & 0.73 & 0.93 & 0.00 & 0.45 & 0.639 & 0.018 & 3 \\
    37 & GLM 5.2 & Open weights & 100 & 0.542 & 27 & 0.82 & 0.79 & 0.94 & 0.00 & 0.46 & 0.637 & 0.020 & 4 \\
    38 & Gemini 3.6 Flash$^{\ast}$ & Google & 100 & 0.593 &  & 0.84 & 0.77 & 0.93 & 0.00 & 0.48 & 0.637 & 0.030 & 4 \\
    39 & Mistral Large 3 & Open weights & 100 & 0.537 & 30 & 0.62 & 0.70 & 0.96 & 0.34 & 0.38 & 0.626 & 0.050 & 5 \\
    40 & GPT-6 Luna$^{\ast}$ & OpenAI & 100 & 0.566 &  & 0.69 & 0.57 & 0.92 & 0.46 & 0.32 & 0.621 & 0.035 & 2 \\
    41 & Grok 4.7$^{\ast}$ & xAI & 99 & 0.556 &  & 0.72 & 0.65 & 0.99 & 0.16 & 0.37 & 0.610 & 0.031 & 0 \\
    42 & GPT-5.5 & OpenAI & 100 & 0.590 & 23 & 0.72 & 0.64 & 0.97 & 0.22 & 0.30 & 0.609 & 0.043 & 1 \\
    43 & Claude Opus 5.5$^{\ast}$ & Anthropic & 100 & 0.515 &  & 0.70 & 0.66 & 0.99 & 0.20 & 0.29 & 0.606 & 0.036 & 2 \\
    44 & GPT-5.4 & OpenAI & 100 & 0.530 & 31 & 0.79 & 0.71 & 0.96 & 0.00 & 0.40 & 0.606 & 0.016 & 1 \\
    45 & Nova 2 Lite & Amazon & 100 & 0.558 & 26 & 0.71 & 0.54 & 0.96 & 0.27 & 0.45 & 0.605 & 0.036 & 2 \\
    46 & Claude Sonnet 4.6 & Anthropic & 100 & 0.521 & 33 & 0.75 & 0.67 & 0.96 & 0.00 & 0.41 & 0.586 & 0.015 & 2 \\
    47 & GPT-6.1 Sol$^{\ast}$ & OpenAI & 100 & 0.479 &  & 0.69 & 0.45 & 0.85 & 0.11 & 0.33 & 0.510 & 0.030 & 2 \\
    \bottomrule
  \end{tabular}}
\end{table}

\section{Reference sample}
\label{app:reference}
Table~\ref{tab:reference} scores the main sources against the full reference of 263 people and against the 100 headline people alone. Section~\ref{sec:res-field} discusses it.

\begin{table}[!htbp]
  \centering\small
  \caption{PSS against two reference samples. \emph{All 263} is the reference of the main text. \emph{Headline 100} scores every source against the real answers of the 100 people whose profiles were simulated. The 163 training people of the out-of-sample copula are not part of this reference, while the in-sample copula is fit on all 263 people and still overlaps with it. The second human sample draws 100 of the 163 people outside the headline sample. The 2000-row copula repeats each headline profile 20 times and shows the copula without its own sampling noise. D, C, M and G are the distribution, correlation, mediation and demographic components against the headline 100.}
  \label{tab:reference}
  \resizebox{\linewidth}{!}{%
  \begin{tabular}{lrrrrrr}
    \toprule
    & All 263 & \multicolumn{5}{c}{Headline 100} \\
    \cmidrule(lr){2-2}\cmidrule(lr){3-7}
    Source & PSS & D & C & M & G & PSS \\
    \midrule
    Second human sample (mean of 20) & 0.92 & 0.99 & 0.73 & 0.56 & 0.53 & 0.79 \\
    Copula, out-of-sample (reported draw) & 0.82 & 0.97 & 0.61 & 0.43 & 0.32 & 0.71 \\
    Copula, out-of-sample (mean of 20) & 0.86 & 0.97 & 0.67 & 0.56 & 0.35 & 0.75 \\
    Copula, out-of-sample (2000 rows) & 0.89 & 0.98 & 0.77 & 0.53 & 0.30 & 0.77 \\
    Copula, in-sample & 0.93 & 0.99 & 0.83 & 0.67 & 0.49 & 0.83 \\
    Best generative (MiniMax M2.7) & 0.81 & 0.82 & 0.74 & 0.95 & 0.65 & 0.84 \\
    gpt-5.4-mini & 0.74 & 0.65 & 0.52 & 0.64 & 0.62 & 0.67 \\
    Jev 1.13, R & 0.68 & 0.76 & 0.64 & 0.79 & 0.51 & 0.75 \\
    Jev 1.13, A & 0.79 & 0.77 & 0.63 & 0.68 & 0.54 & 0.73 \\
    Median generative LLM & 0.70 & 0.74 & 0.70 & 0.71 & 0.55 & 0.75 \\
    Median decision model, R & 0.59 & 0.68 & 0.41 & 0.57 & 0.50 & 0.61 \\
    \bottomrule
  \end{tabular}}
\end{table}

\section{Agreement between models and sensitivity to the weights}
\label{app:agreement}
\begin{figure}[h]
  \centering
  \includegraphics[width=0.85\linewidth]{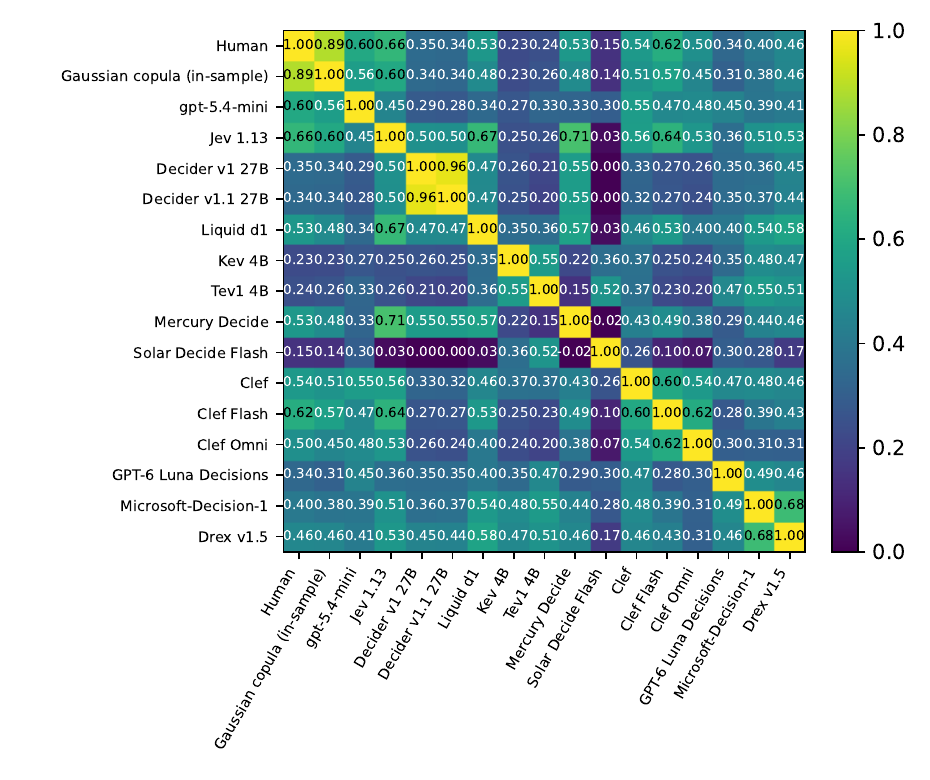}
  \caption{Pearson correlation between the upper triangles of the $68\times68$ item correlation matrices (aligned, rounded readout). The human matrix uses all 263 respondents, and the copula is the in-sample copula. Every other source uses the same 100 profiles.}
  \label{fig:similarity}
\end{figure}

The published version of the original audit found that LLMs agree with each other more than with humans. On the correlation structure, decision models agree with each other (\vSimDecDec) slightly less than with humans (\vSimDecHuman, Figure~\ref{fig:similarity}). Their errors thus show little shared bias on this measure.

The ranking of decision models and baselines is stable under changes of the PSS weights. Over 500 Dirichlet draws the mean Spearman correlation with the canonical ranking is \vWeightRho{} (minimum \vWeightRhoMin; \vWeightFrac{} of draws above 0.95). The in-sample copula is first in every draw. Independent marginals, Tev1 4B and Solar Decide Flash form the bottom three in a share of \vWeightBottomFrac{} of the draws. The changes happen inside the close middle group.

\section{Reproducibility}
\label{app:repro}
All scripts are in \texttt{scripts/psychometry/}. \texttt{run\_decision\_c3.py} sends the decision requests and stores one JSON file per respondent and model with the full probability vectors, the resolved model version, the token count and the cost. It skips cells that are already complete, so an interrupted run can be resumed. \texttt{01\_generate\_synthetic.py} is the generator of the original audit; the new generative models were run with \texttt{--model-set update\_2026\_10\_models --conditions C3\_full --reasoning-effort minimal}. \texttt{score\_decision\_c3.py} computes raw and aligned PSS and recomputes gpt-5.4-mini from the data of the original audit as a check. \texttt{analyze\_decision\_c3.py} runs the readout, response-style, predictive-validity, null-mediation, probability and similarity analyses. \texttt{analyze\_generative\_c3.py} rescores every generative cell with both scoring paths and computes its style and downstream measures. \texttt{bootstrap\_decision\_c3.py} adds the oracle and the out-of-sample copula and runs the paired bootstrap over all sources. \texttt{run\_decision\_controls.py} sends the C0, retest and swap grids, and \texttt{analyze\_decision\_controls.py} analyses them. \texttt{analyze\_decision\_forecasts.py} scores the probability vectors, the references and the generative repeat distributions as forecasts. \texttt{analyze\_decision\_references.py} reads the profile regression as a synthetic respondent and runs the checks against the 100 headline people. \texttt{analyze\_decision\_structure.py} computes the correlation level, the eigenvalue shares, the subscale separation and Tucker's $\varphi$. \texttt{make\_decision\_paper\_assets.py} builds every table, figure and number in this paper from the resulting CSV files. No API call is needed after the generation scripts.

\end{document}